\documentclass[runningheads]{llncs}

\usepackage{times}
\usepackage{soul}
\usepackage{url}
\usepackage[hidelinks]{hyperref}
\usepackage[utf8]{inputenc}
\usepackage[small]{caption}
\usepackage{graphicx}
\usepackage{amsmath}
\usepackage{booktabs}
\usepackage{algorithm}
\usepackage{arevmath}     
\usepackage[noend]{algpseudocode}
\usepackage{listings,multicol}
\usepackage{multicol}
\usepackage{longtable}

\usepackage[caption=false]{subfig}

\usepackage{oz}
\usepackage{makeidx}
\usepackage{listings}
\usepackage{amsfonts}
\usepackage{relsize}
\usepackage{urwchancal}
\usepackage{stmaryrd}
\usepackage{ dsfont }
\usepackage{placeins}
\usepackage{longtable}
\usepackage{standalone}

\usepackage{float}
\newfloat{algorithm}{t}{lop}

\newenvironment{Proof}{\noindent{\bf Proof:}}{$\QED$}

\newtheorem{innercustomthm}{Theorem}
\newenvironment{customthm}[1]
  {\renewcommand\theinnercustomthm{#1}\innercustomthm}
  {\endinnercustomthm}
  
\newtheorem{thm}{Theorem}
\newtheorem{lem}[thm]{Lemma}

\newcommand{\uri}[1]{\ensuremath{{:}{\texttt{#1}}}}  
\newcommand{\urish}[1]{\ensuremath{{\texttt{sh}}{:}{\texttt{#1}}}}
\newcommand{\urisn}[1]{\ensuremath{{\texttt{sn}}{:}{\texttt{#1}}}}
\newcommand{\urirdf}[1]{\ensuremath{{\texttt{rdf}}{:}{\texttt{#1}}}}
\newcommand{\urilam}{\ensuremath{{:}{\lambda}}}
\newcommand{\var}[1]{\ensuremath{\texttt{?#1}}}
\newcommand{\lit}[1]{\ensuremath{\texttt{"#1"}}}
\newcommand{\trip}[3]{\ensuremath{\langle #1,\allowbreak #2,\allowbreak #3 \rangle}}
\newcommand{\critical}{\ensuremath{\texttt{critical}}}
\newcommand{\score}{\ensuremath{\texttt{score}}}

\newcommand{\gpattern}[3]{\ensuremath{[ #1,\allowbreak #2,\allowbreak #3 ]}}

\newcommand{\externalimplementationlink}{\url{https://github.com/paolo7/ISWC2019-code}}

\newcommand{\Nto}{\ensuremath{t^S}} 
\newcommand{\Ntoc}{\ensuremath{t^I}} 
\newcommand{\NtoPrime}{\ensuremath{t_{\score}}} 
\newcommand{\Nta}{\ensuremath{t_A^I}} 
\newcommand{\Ntao}{\ensuremath{t_A^S}} 
\newcommand{\Nts}{\ensuremath{t_A^{\mathds{S}}}}
\newcommand{\NtAPrime}{\ensuremath{t_A^{\prime}}} 
\newcommand{\NtAoPrime}{\ensuremath{t_A^{S\prime}}}
\newcommand{\Ntc}{\ensuremath{t_C}} 
\newcommand{\NtA}{\ensuremath{t_A}} 
\newcommand{\NtAtwo}{\ensuremath{t_A}} 
\newcommand{\Ntq}{\ensuremath{t_q}} 

\usepackage[misc,geometry]{ifsym} 

\hypersetup{
     colorlinks = true,
     linkcolor = blue,
     anchorcolor = blue,
     citecolor = blue,
     filecolor = blue,
     urlcolor = blue
     }

\let\oldmaketitle\maketitle
\renewcommand{\maketitle}{\oldmaketitle\setcounter{footnote}{0}}

\begin{document}

\title{SHACL Constraints with Inference Rules}
\author{Paolo Pareti\ensuremath{^{(\textrm{\Letter})}} \inst{1} \and George	Konstantinidis \inst{1} \and Timothy J.\ Norman \inst{1} \and Murat	\c{S}ensoy \inst{2} }

\authorrunning{P. Pareti et al.}

\institute{University of Southampton, Southampton, United Kingdom \\ \email{pp1v17@soton.ac.uk} \and \"{O}zye\u{g}in University, Istanbul, Turkey}

\maketitle

\begin{abstract}

The Shapes Constraint Language (SHACL) has been recently introduced as a W3C recommendation to define constraints that can be validated against \texttt{RDF} graphs. Interactions of SHACL with other Semantic Web technologies, such as ontologies or reasoners, is a matter of ongoing research. In this paper we study the interaction of a subset of SHACL with inference rules expressed in datalog. On the one hand, SHACL constraints can be used to define a ``schema'' for graph datasets. On the other hand, inference rules can lead to the discovery of new facts that do not match the original schema. Given a set of SHACL constraints and a set of datalog rules, we present a method to detect which constraints could be violated by the application of the inference rules on some graph instance of the schema, and update the original schema, i.e, the set of SHACL constraints, in order to capture the new facts that can be inferred. We provide theoretical and experimental results of the various components of our approach.

\end{abstract}

\section{Introduction}
Information about the type of data contained in a dataset is a critical piece of information both to understand data, and to interface with databases. While the relational model explicitly defines a schema, graph data representations are inherently schemaless, in the sense that any \texttt{RDF} triple could in principle be stored in any \texttt{RDF} triplestore. The Shapes Constraint Language (SHACL) \cite{2017SHACL}, is a W3C recommendation recently introduced to define properties of \texttt{RDF} datasets. SHACL allows the definition of constraints that can be validated against \texttt{RDF} graphs. Such constraints can be seen as the schema of the graphs that do not violate them.

Schemas are not static objects, and they can evolve over time to reflect changes in the datasets they model. One important source of change in graph datasets comes from the application of inference rules. Inference rules can be used to reason about ontological properties, such as class membership. They can also be used for non-ontological types of inference, such as aggregating sensor data to detect facts such as the presence of a fire, or the temperature of a room. This paper focuses on datalog rules \cite{Ceri1989Datalog} without negation, which consist of a head, or consequence, which is instantiated whenever the body, or antecedent is evaluated on the dataset. The exact subset of datalog that we consider is defined in Section \ref{background}.

The application of inference rules might generate new facts, not captured by the original schema definition. Given a set of SHACL constraints and a set of inference rules, we would like to determine whether a graph, initially valid with respect to the SHACL constraints, remains to be valid after computing its closure with the inference rules. If constraint violations can occur, a domain expert could decide whether to remove the inference rules that cause these violations, or to update the violated SHACL constraints. Updating the violated constraints to account for the new facts that can be produced via inferences effectively creates a new schema.

This research is motivated by use cases in the area of Occupational Health and Safety (OHS), and in particular in the mining sector. In these areas, schemas are used to model and understand the underlying data sources, and to ensure interoperability between different tools and applications. Inference rules, usually developed separately, are used to aggregate raw sensor data into more useful abstractions and encode OHS policies (e.g. to detect unsafe working environments). At the moment, domain experts are needed to define such inference rules, however this process is slow, expensive and error prone. Our research aims to better inform experts about the effects of the application of certain rules (which could affect the schema, and therefore interoperability) and automatically detect conflicts between rules and schemas. For example, as schemas frequently change (e.g. sensors malfunctioning, or new ones deployed), it is essential to automatically detect schema changes that render important rules (and policies) no longer applicable, on unforeseen datasets.

Inspired by SHACL constraints, in this paper we present an approach that models \emph{triplestore schemas} as triplets of sets: a set of triple patterns that can be appropriately instantiated by RDF triples, a set of positions in those triples that cannot be instantiated by literal values (e.g., object positions in triples), and a set of existential validity rules (such as tuple-generating dependencies~\cite{fagin2005data}) which must hold on the instatiated triples in order for our graph to be valid. Our triplestore schema captures a fragment of SHACL, but abstracts away from its particular syntax and can be used as a logical tool to model properties of RDF graphs in general. However, it is not meant to provide a complete formal semantic representation of the core SHACL components, such as the one presented in \cite{Corman2018SHACL} to handle recursive SHACL constraints.

Furthermore, we investigate how our triplestore schemas interact with inference rules and evolve into new schemas, that we call \emph{schema consequences}; these are schemas that model all possible RDF graphs extended with the inferred facts. Given an input schema $S$, we want to reason about the applicability of inference rules on all potential instances of $S$, and compute the schema consequence. This problem proves challenging even without taking existential validity rules into account in our schemas, i.e., for what we call our \emph{simple} schema consequence. To reason with inference rules in this version of the problem we have to make use of the notion of a ``canonical'' instance of $S$, representative of all other instances. For this, we first explore such an instance known as the \emph{critical instance} and investigated in relational databases \cite{marnette2009generalized}; running the inference rules on this graph helps us indeed produce our schema consequence. However, the critical instance has a very large and inefficient size and so we turn our attention to finding a much smaller representative instance, that we call the \emph{sandbox} graph. We then present a novel query rewriting algorithm that can compute the simple schema consequence on the sandbox graph, much more efficiently than in the critical instance case.

Building on top of our simple schema consequence we use a novel combination of techniques, variations of datalog rewriting \cite{abiteboulbook} and the chase algorithm \cite{benedikt2017benchmarking}, to produce our \emph{existential-preserving} schema consequence, a triplestore schema the identifies and removes from its description the existential validity rules that could potentially be violated on some instance produced by the inference rules. We provide both theoretical and experimental evaluations of our approaches.

\section{Background} \label{background}

We consider triplestores containing a single \texttt{RDF} \emph{graph}. Such a graph is a set of \emph{triples} $\mathds{U} \times \mathds{U} \times (\mathds{U} \cup \mathds{L})$ where $\mathds{U}$ is the set of all \texttt{IRI}s, $\mathds{L}$ the set of all literals and $\mathds{V}$ the set of all variables. Although we do not explicitly discuss blank nodes, it should be noted that, for the purpose of this paper, they can be treated exactly as \texttt{IRI}s when occurring in a graph. We use the term \emph{constants} to refer to both literals and \texttt{IRI}s. A \emph{graph pattern} is a set of \emph{triple patterns} defined in: $(\mathds{U} \cup \mathds{V}) \times (\mathds{U} \cup \mathds{V}) \times (\mathds{U} \cup \mathds{L} \cup \mathds{V})$. 
Given a pattern $P$, $vars(P)$ and $const(P)$ are the sets of variables and constants in the elements of $P$, respectively. We represent \texttt{IRI}s as namespace-prefixed strings of characters, where a namespace prefix is a sequence of zero or more characters followed by a colon e.g.\ \uri{a}; literals as strings of characters enclosed in double-quotes, e.g.\ \lit{l}, and variables as strings of characters prefixed by a question-mark, e.g.\ \var{v}. The first, second and third elements of a triple $t$ are called, respectively, \emph{subject}, \emph{predicate} and \emph{object}, and are denoted by $t[x]$, $x \in \tau$ with $\tau = \{1,2,3\}$ throughout the paper. 

A \emph{variable substitution} is a partial function $\mathds{V} \pfun \mathds{V} \cup \mathds{U} \cup \mathds{L}$.
A \emph{mapping} is a variable substitution defined as $\mathds{V} \pfun \mathds{U} \cup \mathds{L}$. Given a mapping $m$, if $m(?v) = n$, then we say $m$ contains \emph{binding} $?v \to n$. The domain of a mapping $m$ is the set of variables $dom(m)$. Given a triple or a graph pattern $p$ and a variable substitution $m$ we abuse notation and denote by $m(p)$ the pattern generated by substituting every occurrence of a variable $?v$ in $p$ with $m(?v)$ if $?v\in dom(m)$ (otherwise $?v$ remains unchanged in $m(p)$). A \emph{grounding} is a mapping that transforms a graph pattern into a graph. 

Given a graph pattern $P$ and a graph $I$, we denote the \texttt{SPARQL} evaluation of $P$ over $I$ as the set of mappings $\llbracket P\rrbracket_I$, as defined in \cite{Perez2006Semantics}. 
 A graph pattern \emph{matches} a graph if its evaluation on the graph returns a non-empty set of mappings. 
We consider inference rules $A \rightarrow C$, where $A$ and $C$ are graph patterns, and can be expressed as \texttt{SPARQL} \texttt{construct} queries. Note that essentially both $A$ and $C$ in an inference rule are conjunctive queries~\cite{abiteboulbook}. The \emph{consequent} $C$ of the rule is represented in the \texttt{construct} clause of the query, which is instantiated using the bindings obtained by evaluating the \emph{antecedent} $A$, expressed in the \texttt{where} clause. For technical reasons, we restrict the subset of datalog that we consider with the following requirements: each triple pattern in the consequent $C$ of a rule (1) has a constant in the predicate position, and (2) does not have the same variable in the subject and object position. A single application of an inference rule $r: A \rightarrow C$ to a dataset $I$, denoted by $r(I)$, is $I \cup \bigcup_{m \in \llbracket A \rrbracket_I} \left \{ m(C), \; \text{if} \; m(C)\;  \text{is a valid \texttt{RDF} a graph} \right \} $. 
These rules capture datalog~\cite{abiteboulbook} and subsets of rule notations such as \texttt{SPIN} and \texttt{SWRL} can be represented in this format \cite{Nick2018SWRL2SPIN}. The closure of a dataset $I$ under a set of inference rules $R$, denoted by $clos(I,R)$, is the unique dataset obtained by repeatedly applying all the rules in $R$ until no new statement is inferred, that is,  $clos(I,R) = \bigcup_{i=0}^{i=\infty} I_i$, with  $I_0 = I$, and $I_{i+1} = \bigcup_{r\in R}\{r(I_i)\}$. 

The Shapes Constraint Language (SHACL) defines constraints that can be validated against \texttt{RDF} graphs. An example of a constraint is the requirement for an \texttt{RDF} term to be an \texttt{IRI}. The nodes of an \texttt{RDF} graph against which such constraints are validated are called \emph{focus nodes}. At the core of the SHACL language is the concept of \emph{shapes}. A shape groups together a set of constraints, and defines which focus nodes it should apply to. A shape could either directly target specific specific nodes, such as all the elements of a class, or it could referenced by other shapes. For example, it is possible to define the shape of a ``well-formed email address'', and then specify that every entity of type ``person'' must have at least one email address that satisfies such shape. In this paper we prefix SHACL terms with the namespace \urish{}. 

Given a schema $S$, we denote with $\mathds{I}(S)$ the set of instances of $S$, which are the graphs that $S$ models. We say that two schemas $S$ and $S'$ are semantically equivalent if they model the same set of instances (formally, if $\mathds{I}(S) = \mathds{I}(S')$). Naturally, the interpretation of SHACL constraints as a schema is based on SHACL validation. We say that a graph is an instance of a SHACL schema, defined by its set of constraints, if the graph does not violate the SHACL constraints. Instances of our \emph{triplestore schemas}, instead, are defined in Section~\ref{sec:probDescrExample}. 

\section{Problem Definition}
\label{sec:probdef}

In this section we are going to present our definition of a \emph{triplestore schema}, a simple representation that captures a powerful fragment of SHACL. A set of SHACL shapes $S$ belongs to this fragment if and only if there exists a triplestore schema $S'$ such that $\mathds{I}(S) = \mathds{I}(S')$ (the set of instances of a triplestore schema will be defined later in this section). In essence, triplestore schemas are composed by a set of abstract triple patterns, that intend to model all different triples/instantiations of those patterns, and a set of existential validity constraints that represent ``if-then'' statements of SHACL shapes. Abstracting away from the particulars of the SHACL syntax, on one hand, simplifies our approach and, on the other hand, makes it applicable to other languages as well (e.g., ShEx \cite{Prudhommeaux2014ShapeExpressions}), as long as such a language can be converted into our triplestore schema. Once we have a triplestore schema in place, we define our problem of how do instances of such a schema interact with a set of inference rules, captured in datalog. In particular we would like to reason on schema level and decide if there is a potential instance graph of our schema on which our datalog rules would infer facts that violate the validity constraints.

\subsection{From SHACL to Triplestore Schemas}
\label{sec:probDescrExample}

Our work is inspired by Internet of Things (IoT) settings and as our running example we can consider a dataset of a mining company. The SHACL schema, $S_1$, for this mine is presented in Listing 1. Our mine repository collects data from sensors carried by workers and  deployed in the mine. Data is modelled according to the Semantic Sensor Network Ontology (SSN) \cite{Taylor2017SSN}, with namespace prefix \texttt{s$:$}. In SSN, sensor measurements are called \emph{observations}. The \emph{result} of an observation (e.g.\ ``20'')
relates to a particular \emph{observed property} (e.g.\ temperature) of a particular \emph{feature of interest} (e.g.\ a room). In our example the mine contains two types of sensors. The first is a carbon monoxide (CO) detector, which records value ``0'' if the CO concentration is within the allowed limits, and ``1'' otherwise.  The second is an RFID reader used to locate personnel in the mine by sensing the nearby RFID tags carried by the mine workers. SHACL shape \uri{s0} specifies that the collected sensor data will only refer to those two sensor types. The dataset of the mine is expected to contain a list of known personnel RFID tags, and information on who is currently carrying them. Shape \uri{s1} specifies that for every personnel tag, we know who it is carried by. Shapes \uri{s2} and \uri{s3} restrict features of interest to being \texttt{IRI}s and measurement results to be \texttt{IRI}s or literals. Shape \uri{s4} declares that the sensor data contains instances of only two classes, namely sensor observations, and personnel tags. The following listing describes schema $S_1$:

\footnotesize \begin{lstlisting}
:s0   a                   sh:NodeShape ;
      sh:targetObjectsOf  sn:observedProperty ;
      sh:in               ( :COLevel :TagID ) .
:s1   a                   sh:NodeShape ;
      sh:targetClass      :PersonnelTag ;
      sh:property         [ sh:minCount  1 ;
                            sh:path      :carriedBy ] .
:s2   a                   sh:NodeShape ;
      sh:targetObjectsOf  sn:hasFeatureOfInterest ;
      sh:nodeKind         sh:IRI .
:s3   a                   sh:NodeShape ;
      sh:targetObjectsOf  sn:hasResult ;
      sh:nodeKind         sh:IRIOrLiteral .
:s4   a                   sh:NodeShape ;
      sh:targetObjectsOf  rdf:type ;
      sh:in               ( sn:Observation :PersonnelTag ) .
\end{lstlisting} \normalsize

When using SHACL as a schema language, we would like the constraints to describe the type of data contained in a dataset as accurately as possible. 
SHACL constraints usually target only a limited number of predicates in an RDF graph, and triples with predicates other than the targeted ones could be present in the graph without causing violations. However, for our purposes we adopt a closed-world view of the available vocabulary of predicates, and we would like to restrict valid graphs to only contain a fixed set of predicates. This vocabulary restriction can be specified by an appropriate SHACL constraint that uses the \urish{closed} component. Hence, subsequently we assume that all SHACL schemas that we work with contain such a component which specifies that valid instances of this schema do not contain predicates other than the ones that appear in the SHACL shapes. 
This is inline with relational databases where the discovery of completely new types of facts, i.e., triples with unforseen predicates, would be reflected by a corresponding change in the original schema.

\begin{figure}
\begin{multicols}{2}
\footnotesize \begin{lstlisting}
$\var{v1}\;$sn:observedProperty$\;$:COLevel$\,$. 
$\var{v2}\;$sn:observedProperty$\;$:TagID$\,$.
$\var{v3}\;$rdf:type$\;$sn:Observation$\,$.
$\var{v4}\;$rdf:type$\;$:PersonnelTag$\,$.
$\var{v5}\;$:carriedBy$\;$?v6$\,$.
$\var{v7}\;$sn:hasFeatureOfInterest$\;$?v8$\,$.
$\var{v9}\;$sn:hasResult$\;$?v10$\,$.
\end{lstlisting} \normalsize
\end{multicols}
\caption{Graph pattern $S_1^G$.}
\label{fig:graphPatternSG1}
\end{figure}

In our running example, instances of schema $S_1$ would contain some triples of the form of the triples patterns of graph pattern $S_1^G$ displayed in Fig.\ \ref{fig:graphPatternSG1}, where each variable can be appropriately instantiated by an IRI or a literal. In fact, exactly such a set of triple patterns will be the first element of our representation of triplestore schemas, called a \emph{shema graph}, as defined subsequently. Note that valid instances of our schema might contain multiple instantiations of some, but not necessarily all of the predicates defined in the schema graph, and they cannot contain other kinds of triples (i.e., undefined predicates). We use different variables in $S_1^G$ to denote the fact that variables in a schema graph act as wildcards, and are not meant to join triple patterns together.

In addition to the schema graph, a second part of our schema representation will be the subset of variables from the schema graph, called the \emph{no-literal set}, where literals can not occur in valid instances. For example, we cannot instantiate variables \var{v7} and \var{v8} of triple pattern  $\gpattern{\var{v7}}{\urisn{hasFeatureOfInterest}}{\var{v8}}$ from Fig.\ \ref{fig:graphPatternSG1} with a literal; in the case of \var{v7}, because we would not generate a valid \texttt{RDF} triple, and in the case of \var{v8}, because it would violate shape \uri{s2}.

The last part of our schema representation will translate SHACL constraints to ``if-then'' statements, like the following which corresponds to shape \uri{s1} of schema $S_1$:
\footnotesize \begin{lstlisting}
$e_1$ = $ \gpattern{\var{v1}}{\urirdf{type}}{\uri{PersonnelTag}}$ $\rightarrow^{\exists}$ $ \gpattern{\var{v1}}{\uri{carriedBy}}{\var{v2}}$
\end{lstlisting} \normalsize

These constraints are essentially \emph{existential rules}~\cite{baget2011rules}, also expressible as tuple-generating dependencies (TGDs)~\cite{fagin2005data}. For all practical purposes, the part of SHACL that we consider, when translatable to existential rules, falls in a language known as linear weakly-acyclic TGDs~\cite{fagin2005data} with a single atom in the consequent. Linear means that these rules have only one atom, i.e., one triple pattern, in the antecedent, and \emph{weakly-acyclic} is a property that guarantees that forward-chaining algorithms, such as the Chase~\cite{benedikt2017benchmarking} algorithms terminate when run with these rules. 

In addition, in this paper, we only deal with atomic consequents as well.  
Formally, an existential rule is defined as a formula of the form: 
$a \rightarrow^{\exists} c$, where $a$ and $c$, respectively the antecedent and the consequent of the rule, are triple patterns. 
The consequent specifies which triples must exist in a graph whenever the antecedent holds in that graph. We say that an existential rule $a \rightarrow^{\exists} c$ is \emph{violated} on a graph $I$ if there exists a mapping $m \in \llbracket \{a\}\rrbracket_I$ such that $\llbracket m(c) \rrbracket_I = \emptyset$ (i.e., $m(c)$ is not in $I$); the rule is \emph{satisfied} otherwise. Note that if $m(c)$ is a ground triple, $\llbracket m(c) \rrbracket_I$ is not empty, as it contains the empty mapping \cite{Perez2006Semantics}. Given a set of existential rules $E$, we denote with $violations(E,I)$ the set of pairs $\langle m,e \rangle$, where $e \in E$ and mapping $m$ causes $e$ to be violated on instance $I$.

We are now ready to define our triplestore schemas. A triplestore schema (or from now on, just \emph{schema}) $S$, is a tuple $\langle S^G, S^\Delta, S^{\exists} \rangle$, where $S^G$, called a \emph{schema graph}, is a set of triple patterns where every variable occurs at most once, $S^\Delta$ is a subset of the variables in $S^G$ which we call the \emph{no-literal} set, and $S^{\exists}$ is a set of existential rules. Intuitively, $S^G$ defines the type of triples that can appear in a graph, where variables act as wildcards, which can be instantiated with any constant element. To account for the restrictions imposed by the \texttt{RDF} data model, the no-literal set $S^\Delta$ defines which variables cannot be instantiated with literals, thus $S^\Delta$ must at least include all variables that occur in the subject or predicate position in $S^G$. For example, if $\trip{\var{v1}}{\urisn{hasResult}}{\var{v2}} \in S^G_{'}$ and $\var{v2} \not \in S^\Delta_{'}$, then the instances of schema $S_{'}$ can contain any triple that has \urisn{hasResult} as a predicate. If $\trip{\var{v3}}{\urirdf{type}}{\uri{Observation}} \in S^G_{'}$ and $\var{v3} \in S^\Delta_{'}$, the instances of $S_{'}$ can contain any entity of type \uri{Observation}. While $S^G$ and $S^\Delta$ together define the set of all the possible triples that can be found in a graph, not all combinations of such triples are valid instances of the schema. The set of existential rules $S^{\exists}$ defines further requirements that instances of the schema must satisfy. Formally, a graph $I$ is an \emph{instance} of a triplestore schema $\langle S^G, S^\Delta, S^{\exists} \rangle$ if and only if $violations(S^{\exists}, I) = \emptyset$ and for every triple $t^I$ in $I$ there exists a triple pattern $t^S$ in $S^G$, such that $t^I$ is an \emph{instantiation} of $t^S$ w.r.t to $S^{\Delta}$, that is, there exists a mapping $m$ such that (1) $m(t^S) = t^I$ and (2) $m$ does not bind any variable in $S^\Delta$ to a literal. 

For our SHACL to triplestore schema translation we direct the reader to our appendix and our implementation in our code repository.\footnote{\label{externalImplementationLink}\externalimplementationlink}

\subsection{Inference Rules and Schema consequences}
\label{sec:RulesAndConsequences}

In this paper we are interested in the effect inference rules, not to be confused with existential rules, have on RDF graphs, and their interaction with existential rules. Inference rules are used to compute the \emph{closure} of instances of our original schema as defined in Section~\ref{background}. As an example consider the of inference rules $R_1 =\{ r_1$, $r_2$, $r_3\}$ below. Rule $r_1$ states that the RFIDs recorded by the sensors should be interpreted as personnel tags, and it records the location of where they are detected. Rule $r_2$ states that locations with a high carbon monoxide (CO) concentration should be off-limit. Rule $r_3$ states that if someone is located in an off-limit area, then they are trespassing in that area.

\footnotesize \begin{lstlisting}
$r_1$ = $\{\:\,\gpattern{\var{v1}}{\urisn{observedProperty}}{\uri{TagID}},$
      $\gpattern{\var{v1}}{\urisn{hasResult}}{\var{v2}},$     
      $\gpattern{\var{v1}}{\urisn{hasFeatureOfInterest}}{\var{v3}}\:\}$
      $\rightarrow$ $\{\:\gpattern{\var{v2}}{\urirdf{type}}{\uri{PersonnelTag}},
      \gpattern{\var{v2}}{\uri{isLocatedIn}}{\var{v3}}\:\}$
$r_2$ = $\{\:\,\gpattern{\var{v1}}{\urisn{observedProperty}}{\uri{COLevel}},$
      $\gpattern{\var{v1}}{\urisn{hasResult}}{\lit{1}},$
      $\gpattern{\var{v1}}{\urisn{hasFeatureOfInterest}}{\var{v2}}\:\}$
      $\rightarrow$ $\{\:\gpattern{\var{v2}}{\urirdf{type}}{\uri{OffLimitArea}}\:\}$
$r_3$ = $\{\:\,\gpattern{\var{v1}}{\uri{isLocatedIn}}{\var{v2}},$
      $\gpattern{\var{v2}}{\urirdf{type}}{\uri{OffLimitArea}}\:\}$
      $\rightarrow$ $\{\:\gpattern{\var{v1}}{\uri{isTrespassingIn}}{\var{v2}}\:\}$
\end{lstlisting} \normalsize

In our example, an emergency response application  might need to know who is carrying each personnel RFID tag, in order to compute an emergency response plan. In this case, it is important to know which existential rules the application of a set of inference rules can violate. Once potential violations are detected, a domain expert could, for example, decide whether to relax (i.e.\ remove) the violated existential rules, or to remove the inference rules that cause the violations. 

Thus, an example of the central question we address in this paper is: is $e_1$ guaranteed to remain valid in instances of schema $S_1$ under closure with inference rules $R_1$? The answer to this question is \emph{no}, as demonstrated by graph $I_1$, which is a valid instance of $S_1$. This instance contains two records of a miner tag being detected, namely \uri{WID1} and \uri{WID2}. While we know that \uri{WID1} is being carried by worker \uri{Alex}, we do not have any such information about tag \uri{WID2}.

\begin{figure}
\begin{multicols}{2}
\footnotesize \begin{lstlisting}
$:$o1$\;$sn$:$observedProperty$\;:$TagID$\,$;
  sn$:$hasFeatureOfInterest$\;:$room1$\,$;
  sn$:$hasResult$\;$:WID1$\,$.
$:$o2 sn$:$observedProperty$:$TagID$\,$;
  sn$:$hasFeatureOfInterest$:$room2$\,$;
  sn$:$hasResult$\;:$WID2$\,$.
$:$o3 sn$:$observedProperty$\;$$:$COLevel$\,$;
  sn$:$hasFeatureOfInterest$\;$$:$room2$\,$;
  sn$:$hasResult$\;$"1"$\,$.
$:$WID1$\;$a$\;$$:$PersonnelTag$\,$;
  :carriedBy$\;$$:$Alex$\,$.
\end{lstlisting} \normalsize
\end{multicols}
\caption{Instance $I_1$.}
\end{figure}

Rule $r_1$ will deduce that \uri{WID2} is a personnel tag, by inferring triples $\gpattern{\uri{WID2}}{\urirdf{type}}{\uri{PersonnelTag}}$ and $\gpattern{\uri{WID2}}{\uri{isLocatedIn}}{\uri{room2}}$ from instance $I_1$. However, since there is no information on who is carrying tag \uri{WID2}, existential rule $e_1$ is violated. A domain expert analysing this conflict can then decide to either relax  $e_1$, to state that there is not always information on who is carrying a personnel tag, or to remove rule $r_1$, to state that not all RFIDs recorded by the sensors are personnel tags.
Rule $r_2$ is triggered by observation \uri{o3}, inferring triple $\gpattern{\uri{room2}}{\urirdf{type}}{\uri{OffLimitArea}}$. The \texttt{IRI} \uri{OffLimitArea} is not one of the types allowed in the original schema. Therefore, we might want to either revise rule $r_2$, or extend schema $S_1$ to allow for instances of this type.
Facts inferred by rules $r_1$ and $r_2$ together trigger rule $r_3$, which will infer $\gpattern{\uri{WID2}}{\uri{isTrespassingIn}}{\uri{room2}}$, i.e., that the person carrying the RFID tag \uri{WID2} is trespassing in dangerous area \uri{room2}. 
These new facts contain the new predicate \uri{isTrespassingIn}, and thus violate our closed-world interpretation of schema $S_1$ (as captured by our schema graph patterns). Hence, if one want to retain all inference rules $R_1$ in our mine repository a natural alteration of the original schema (and its schema graph) is required.

In this paper we deal with predicting the above mentioned constraint violations, and computing an updated schema that accounts for them, without looking at specific instances such as $I_1$. Given a schema $S : <S^G,S^{\Delta},S^{\exists}>$ and a set of inference rules $R$, we want to compute a new schema, called \emph{schema consequence}, which captures all the inferences of the set of rules $R$ on any potential instance of $S$. 
By computing an updated triplestore schema, once a violation is detected, our approach gives preference to maintaining inference rules over maintaining the original schema, essentially choosing to alter the schema graph and/or the existential rules. This is not an inherent limitation of our approach, which could be easily transformed to a method that maintains the original schema and chooses to reject conflicting inference rules.

To present our problem incrementally, we first compute a \emph{simple} schema consequence which does not take existential rules into account, i.e., it only deals with $S^G$ and $S^
{\Delta}$ of our triplestore schema, and then we extend our solution to take $S^{\exists}$ into account in our \emph{existentially-preserving} schema consequence.

The simple interpretation of a schema consequence captures the type of triples that the closure of an instance of the schema could potentially contain. Given a schema $S$ and a set of inference rules $R$, a schema $S^{\prime}$ is a \emph{simple schema consequence} of $S$ with respect to $R$, denoted $con(S,R)$, if $\mathds{I}(S^{\prime}) = \bigcup_{I \in \mathds{I}(S)} \{I' | I' \subseteq clos(I,R)$\}. It is important to notice that every subset of an instance's closure is still an instance of the simple schema consequence. Thus a simple schema consequence can contain the consequence of an inference rule application without containing a set of triples matching the antecedent, or vice versa. This situation is commonly encountered when some triples are deleted after an inference is made. Effectively, this definition does not assume that all the triples in an instance's closure are retained.
One use of this schema consequence is to discover whether certain important facts (e.g.\ personnel trespassing in a dangerous area) can be inferred from the given schema (e.g.\ the available sensor data streams) and set of inference rules (e.g.\ the sensor data aggregation rules). Another use is to compute which inference rules are \emph{applicable} on a schema, meaning that they will be triggered on at least one instance of that schema. We provide two approaches to compute it in Sec. \ref{sec:approach}.

Given a schema $S$ and a set of inference rules $R$, a schema $S^{\prime}$ is an existential-preserving schema consequence of $S$ with respect to $R$, denoted $con^{ex}(S,R)$, if and only if $\mathds{I}(S^{\prime}) = \bigcup_{I \in \mathds{I}(S)} \{I' | I' \subseteq clos(I,R) \land violations(S^{\exists},I') = violations(S^{\exists},clos(I,R))\}$. 
In other words, instances of an existential-preserving schema consequence are generated by computing the closure of an instance of the original schema under the inference rules, and then discarding a set of triples as long as doing so does not generate new violations of existential rules $S^{\exists}$. This allows us to detect which existential rules can be violated by the application of inference rules (and not just by arbitrary triple deletions).

\section{Computing the Simple Schema Consequence}
\label{sec:approach}

We compute $con(S,R)$ iteratively, on a rule-by-rule basis. In correspondence to a single application $r(I)$, of an inference rule $r$ on an instance $I$, we define a \emph{basic consequence} of a schema $S$ by an inference rule $r$, denoted by $r(S)$, as a finite schema $S^{\prime}$ for which $\mathds{I}(S') = \bigcup_{I \in \mathds{I}(S)} \{I' | I' \subseteq r(I)\}$. It is now easy to see that the consequence schema for a set of inference rules $con(S,R)$ is obtained by repeatedly executing $r(S)$ for all $r \in R$ until no new pattern is inferred. Formally, $con(S,R) = \bigcup_{i=0}^{i=n} S_i$, with  $S_0 = S$, and $S_{i+1} = \bigcup_{r\in R}\{r(S_i)\}$, and $S_n= S_{n-1}$ (modulo variable names). In this section we focus on computing a single basic schema consequence $r(S)$, and describe two approaches for this, namely Schema Consequence by Critical Instance ($\critical(S,r)$), and Schema Consequence by Query Rewriting ($\score(S,r)$). 

Given a schema $S$ and an inference rule $r: A \to C$, our approach to compute the basic schema consequence for $r$ on $S$ is based on evaluating $A$, or an appropriate rewriting thereof, on a ``canonical'' instance of $S$, representative of all instances modelled by the schema. The mappings generated by this evaluation are then (1) filtered (in order to respect certain literal restrictions in \texttt{RDF})  and (2) applied appropriately to the consequent $C$ to compute the basic schema consequence.

We present two approaches, that use two different canonical instances. The first instance is based on the concept of a \emph{critical instance}, which has been investigated in the area of relational databases before~\cite{marnette2009generalized} (and similar notions in the area of Description Logics~\cite{glimm2014abstraction}). Adapted to our \texttt{RDF} setting, the critical instance would be created by substituting the variables in our schema, in all possible ways, with constants chosen from the constants in $S^G$ and $A$ as well as a new fresh constant not in $S^G$ or $A$. In~\cite{marnette2009generalized} this instance is used in order to decide Chase termination; 
Chase is referred to rule inference with \emph{existential} variables, more expressive than the ones considered here and for which the inference might be infinite (see \cite{benedikt2017benchmarking} for an overview of the Chase algorithm). Although deciding termination of rule inference is slightly different to computing the schema consequence, we show how we can actually take advantage of the critical instance in order to solve our problem. Nevertheless, this approach, that we call \critical, creates prohibitively large instances when compared to the input schema. Thus, later on in this section we present a rewriting-based approach, called \score, that runs a rewriting of the inference rule on a much smaller canonical instance of the same size as $S^G$.

\noindent {\bf The Critical Approach.} For both versions of our algorithms we will use a new fresh \texttt{IRI} $\urilam$ such that $\urilam \not\in const(S^G) \cup const(A)$. 
Formally, the critical instance $\mathds{C}(S,A \rightarrow C)$ is the set of triples: 
\[ \{t \lvert\text{ triple }t\text{ with }t[i] =  \left \{ \begin{tabular}{ll} 
$c$ & if $t^S[i]$ is a variable and: 
\\
 & (1) $c$ is a \texttt{IRI} or 
 \\
 & (2) $i=3$ and $t^S[i] \not\in S^\Delta$
\\ 
$t^S[i]$ & if $t^S[i]$ is not a variable
\end{tabular} \right \},\\ t^S \in S^G, i \in \tau, c \in const(S^G) \cup const(A) \cup \{\urilam\} \}\]
The critical instance replaces variables with \texttt{IRI}s and literals from the set $const(S^G) \cup const(A) \cup \{\urilam\}$, while making sure that the result is a valid \texttt{RDF} graph (i.e.\ literals appear only in the object position) and that it is an instance of the original schema (i.e.\ not substituting a variable in $S^\Delta$ with a literal). 
In order to compute the triples of our basic schema consequence for inference rule $r$ we evaluate $A$ on the critical instance, and post-process the mappings $\llbracket A\rrbracket_{\mathds{C}(S,r)}$ as we will explain later. 
Before presenting this post-processing of the mappings we stretch the fact that this approach is inefficient and as our experiments show, non scalable.  For each triple $t$ in the input schema $S$, up to $\lvert const(S^G) \cup const(A) \cup \{\urilam\}\rvert ^{vars(t)}$ new triples might be added to the critical instance.

\noindent{{\bf The Score Approach.}} To tackle the problem above we present an alternative solution based on query rewriting, called \score . This alternative solution uses a small instance called the \emph{sandbox} instance which is obtained by taking all triple patterns of our schema graph $S^G$ and substituting all variables with the same fresh \texttt{IRI} $\urilam$. This results in an instance with the same number of triples as $S^G$. The main property that allows us to perform this simplification is the fact that variables in $S^G$ are effectively independent from each other. Formally, a sandbox graph $\mathds{S}(S)$ is the set of triples:
\[ \{t \lvert\text{ triple }t\text{ with }t[i] =  \left \{ \begin{tabular}{ll} 
$\urilam$ & if $t^S[i]$ is a variable,
\\ 
$t^S[i]$ & else
\end{tabular} \right \}, t^S \in S^G, i \in \tau \}\]
Contrary to the construction of the critical instance, in our sandbox graph, variables are never substituted with literals (we will deal with \texttt{RDF} literal peculiarities in a post-processing step). Also notice that $\mathds{S}(S) \in \mathds{I}(S)$ and $\mathds{S}(S) \subseteq \mathds{C}(S,r)$. 
As an example, consider the sandbox graph $\mathds{S}(S_1)$ of schema $S_1$ from Section \ref{sec:probDescrExample}:

\begin{multicols}{2}
\footnotesize \begin{lstlisting}
$\urilam\;$sn:observedProperty$\;$:COLevel$\;$. 
$\urilam\;$sn:observedProperty$\;$:TagID$\;$.
$\urilam\;$rdf:type$\;$sn:Observation$\;$.
$\urilam\;$rdf:type$\;$:PersonnelTag$\;$.
$\urilam\;$:carriedBy$\;\urilam\;$.
$\urilam\;$sn:hasFeatureOfInterest$\;\urilam\;$.
$\urilam\;$sn:hasResult$\;\urilam\;$.
\end{lstlisting} \normalsize
\end{multicols}

The critical instances $\mathds{C}(S_1,r_1)$, $\mathds{C}(S_1,r_2)$ and $\mathds{C}(S_1,r_3)$ from our example would contain all the triples in $\mathds{S}(S_1)$, plus any other triple obtained by substituting some variables with constants other than $\urilam$. For example, $\mathds{C}(S_1,r_2)$ would contain the triple $\gpattern{\urilam}{\urisn{hasResult}}{\uri{OffLimitArea}}\}$.

In order to account for all mappings produced when evaluating $A$ on $\mathds{C}(S,r)$ we will need to evaluate a different query on our sandbox instance, essentially by appropriately rewriting $A$ into a new query. 
To compute mappings, we consider a rewriting $\mathds{Q}(A)$ of $A$, which expands each triple pattern $t_A$ in $A$ into the union of the $8$ triple patterns that can be generated by substituting any number of elements in $t_A$ with $\urilam$. Formally, $\mathds{Q}(A)$ is the following conjunction of disjunctions of triple patterns, where $\bigwedge$ and $\bigvee$ denote a sequence of conjunctions and disjunctions, respectively:
\begin{equation*}
\mathds{Q}(A) = \bigwedge{}_{t \in A} \Bigg( \bigvee{}_{\substack{x_1 \in \{\urilam,t[1]\} \\ x_2 \in \{\urilam,t[2]\} \\ x_3 \in \{\urilam,t[3]\} }} 
\trip{x_1}{x_2}{x_3} \Bigg)
\end{equation*}

When translating this formula to SPARQL we want to select mappings that contain a binding for all the variables in the query, so we explicitly request all of them in the select clause. 
For example, consider graph pattern $A_1 = \{\trip{\var{v3}}{\uri{a}}{\var{v4}}, \trip{\var{v3}}{\uri{b}}{\uri{c}}\}$, which is interpreted as query:
\footnotesize \begin{lstlisting}
SELECT ?v3 ?v4 WHERE { ?v3 :a ?v4 . ?v3 :b :c }
\end{lstlisting}  \normalsize
Query rewriting $\mathds{Q}(A_1)$ then corresponds to:
\footnotesize \begin{lstlisting}
SELECT ?v3 ?v4 WHERE {
  { {?v3 $:$a ?v4} UNION {$:$$\lambda$ $:$a ?v4} UNION {?v3 $:$$\lambda$ ?v4}
    UNION {?v3 $:$a $:$$\lambda$} UNION {$:$$\lambda$ $:$$\lambda$ ?v4} UNION {$:$$\lambda$ $:$a $:$$\lambda$}
    UNION {?v3 $:$$\lambda$ $:$$\lambda$} UNION {$:$$\lambda$ $:$$\lambda$ $:$$\lambda$} }
  { {?v3 $:$b $:$c} UNION {$:$$\lambda$ $:$b $:$c} UNION {?v3 $:$$\lambda$ $:$c}
    UNION {?v3 $:$b $:$$\lambda$} UNION {$:$$\lambda$ $:$$\lambda$ $:$c} UNION {$:$$\lambda$ $:$b $:$$\lambda$}
    UNION {?v3 $:$$\lambda$ $:$$\lambda$} UNION {$:$$\lambda$ $:$$\lambda$ $:$$\lambda$} } }
\end{lstlisting}  \normalsize

\noindent Below we treat $\mathds{Q}(A)$ as a union of conjunctive queries, or \texttt{UCQ} \cite{abiteboulbook}, and denote  $q \in \mathds{Q}(A)$ a conjunctive query within it.

We should point out that in this section we present a generic formulation of both approaches that is applicable to schema graphs having variables in the predicate position. If variables cannot occur in this position, such as in the triplestore schemas representation of SHACL constraints, these approaches could be optimised; for example by removing from $\mathds{Q}(A)$ all the triples patterns that have \urilam\ in the predicate position.

Having defined how the \critical\ and \score\ approaches compute a set of mappings, we now describe the details of the last two phases required to compute a basic schema consequence.

\noindent {\bf Filtering of the mappings} This phase deals with processing the mappings computed by either  \critical\ or \score, namely $\llbracket A\rrbracket_{\mathds{C}(S,r)}$ or $\llbracket\mathds{Q}(A)\rrbracket_{\mathds{S}(S)}$. It should be noted that it is not possible to simply apply the resulting mappings on the consequent of the inference rule, as such mappings might map a variable in the subject or predicate position to a literal, thus generating an invalid triple pattern. Moreover, it is necessary to determine which variables should be included in the no-literal set of the basic schema consequence. 
The schema $S^{\prime}$, output of our approaches, is initialised with the same graph and no-literal set as $S$, and with an empty set of existential rules. Formally $S^{\prime}$ is initialised to $\langle S^G, S^\Delta, \emptyset \rangle$.  
We then incrementally extend $S^{\prime}$ on a mapping-by-mapping basis until all the mappings have been considered, at which point, $S^{\prime}$ is the final output of our basic schema expansion. 

For each mapping $m$ in $\llbracket A\rrbracket_{\mathds{C}(S,r)}$ or $\llbracket\mathds{Q}(A)\rrbracket_{\mathds{S}(S)}$, we do the following. We create a temporary no-literal set $\Delta^m$. This set will be used to keep track of which variables could not be bound to any literals if we evaluated our rule antecedent $A$ on the instances of $S$, or when instantiating the consequence of the rule. We initialise $\Delta^m$ with the variables of our inference rule $A\rightarrow C$ that occur in the subject or predicate position in some triple of $A$ or $C$, as we know that they cannot be matched to or instantiated with literals. 

Then, we consider the elements that occur in the object position in the triples $t_A$ of $A$. We take all the rewritings $t_q$ of $t_A$ in $\mathds{Q}(A)$ (if using \critical , it would be enough to consider a single rewriting $t_q$ with $t_q = t_A$). Since the mapping $m$ has been computed over the canonical instance ($\mathds{S}(S)$ or $\mathds{C}(S,r)$ depending on the approach), we know that there exists at least one $t_q$ such that $m(t_q)$ belongs to the canonical instance. 
We identify the set of schema triples $t^S \in S$ that model $m(t_q)$, for any of the above $t_q$. Intuitively, these are the schema triples that enable $t_A$, or one of its rewritings, to match the canonical instance with mapping $m$. If $t_A[3]$ is a literal $l$, or a variable mapped to a literal $l$ by $m$, we check if there exists any $t^S$ from the above such that $t^S[3] = l$ or $t^S[3]$ is a variable that allows literals (not in $S^\Delta$). If such triple pattern doesn't exist, then $m(A)$ cannot be an instance of $S$ since it has a literal in an non-allowed position, and therefore we filter out or \emph{disregard} $m$. 
If $t_A[3]$ is a variable mapped to $\urilam$ in $m$, we check whether in any of the above $t^S$, $t^S[3]$ is a variable that allows literals (not in $S^\Delta$). 
If such $t^S$ cannot be found, we add variable $t_A[3]$ to $\Delta^m$. Intuitively, this models the fact that $t_A[3]$ could not have been bound to literal elements under this mapping. 
Having considered all the triples $t_A \in A$ we filter out mapping $m$ if it binds any variable in $\Delta^m$ to a literal. If $m$ is not filtered out, we say that inference rule $r$ is applicable, and we use $m$ to expand $S^{\prime}$.
 
\noindent {\bf Schema Expansion.} 
For each mapping $m$ that is not filtered out, we compute the substitution $s^m$, which contains all the bindings in $m$ that map a variable to a value other than $\urilam$, and for every binding $?v \to \urilam$ in $m$, a variable substitution $?v \to ?v^*$ where $?v^*$ is a fresh new variable. We then add triple patterns $s^m(m(C))$ to $S^{\prime G}$ and then add the variables $s^m(\Delta^m) \cap vars(S^{\prime G})$ to $S^{\prime \Delta}$. 
Although the schema consequences produced by $\score(S,r)$ and $\critical(S,r)$ might not be identical, they are semantically equivalent (i.e.\ they model the same set of instances). This notion of equivalence is captured by the following theorem.
\begin{customthm}{1} \label{theorem:approachEquality}
  For all rules $r: A \rightarrow C$ and triplestore schemas $S$, $\mathds{I}(\score(S,r)) = \\ \mathds{I}(\critical(S,r))$.
\end{customthm}

\noindent The \score\ approach (and by extension also \critical , by Theorem \ref{theorem:approachEquality}) is sound and complete. The following theorem captures the this notion by stating the semantic equivalence of $\score(S,r)$ and $r(S)$. For our proofs, we refer the reader to our appendix. 
\begin{customthm}{2} \label{newTheoremSimple} For all rules $r: A \rightarrow C$ and triplestore schemas $S$, $\mathds{I}(\score(S,r)) = \\ \mathds{I}(r(S))$.

\end{customthm}

\noindent{\bf{Termination.}}  It is easy to see that our approaches terminate since our datalog rules do not contain existential variables, and do not generate new \texttt{IRI}s or literals (but just fresh variable names). After a finite number of iterations, either approach will only generate isomorphic (and thus equivalent) triple patterns.

\section{Computing the Existential-Preserving Schema Consequence}
\label{sec:approachWithExistentials}

In order to compute the existential-preserving schema consequence we are going to build on the result of our simple schema consequence. Recall the definitions of our schema consequences from Section \ref{sec:RulesAndConsequences} and note that given a schema $S = \langle S^G, S^{\Delta},S^{\exists} \rangle$ and a set of inference rules $R$ such that $con(S,R) = \langle S'^G, S'^{\Delta}, \emptyset \rangle$ then $con^{ex}(S,R) = \langle S'^G, S'^{\Delta}, S'^{\exists}\rangle$ for some $ S'^{\exists} \subseteq S^{\exists}$; that is, the output schema graph and no-literal set of the existential-preserving schema consequence are the same as those of the simple one.

Thus, our first step is to compute the schema graph and no-literal set of the existential preserving schema consequence, using Section \ref{sec:approach}. Next, and in the rest of this section, we want to compute the set of existential rules $S'^{\exists}$ that are still valid on all possible ``closure'' instances (instances of the original schema closed under $R$), or complementary, those existential rules that are violated on some ``closure'' instance. 

Starting from an instance $I$ of $S$, which by definition satisfies $S^{\exists}$, an existential rule might become violated by the inference rules due to new facts added by the closure. 
Thus, the algorithm that we are going to present intends to find an instance $I$ of $S$, that can ``trigger'' an existential rule $a \rightarrow^{\exists} c$ by mapping its antecedent $a$ on $clos(I,R)$. For every existential rule, we want to construct $I$ in a ``minimal'' way, so that if $clos(I,R)$ satisfies the rule $e$ then there is no proper subset $I'$ of $I$ which is still an instance of $S$ and does not satisfy the rule. By considering all such minimal instances $I$ for every existential rule, we can determine if the rule is violated or not on any potential closure of an instance.

\begin{algorithm}
\footnotesize
    \caption{Computation of the existential rules in $con^{ex}(S,R)$}
    \label{algExistential}
    \begin{algorithmic}[1]
        \Procedure{retainedExistentials}{$S: \langle S^G, S^{\protect\Delta}, S^{\protect\exists} \rangle, R$}
        \State $V \leftarrow \emptyset$
        \For{each $e: a \rightarrow^{\exists} c \in S^{\exists}$}
                \For{each $r: A \rightarrow C \in R$}
                     \If{$\llbracket \mathds{Q}(a)\rrbracket_{\mathds{S}(C)} \not = \emptyset$}
                        \State $W \leftarrow$ all rewritings of the antecedent of $r: A \rightarrow C$ with rules $R$
                        \For{each $w: A^w \rightarrow C \in W$}
                            \State $M^w \leftarrow \llbracket \mathds{Q}(A^w)\rrbracket_{\mathds{S}(S)}$
                            \For{each $m^w \in M^w$}
                                \State $\tilde{m}^w \leftarrow $ all mappings in $m^w$ that do not map a variable to \urilam
                                \State $g \leftarrow$ mapping from the $vars(\tilde{m}(A^w))$ to fresh \texttt{IRI}s
                                \State $I^g \leftarrow g(\tilde{m}(A^w))$ 
                                \State $I^{g'} \leftarrow \emptyset$ 
                                \While{$I^g \not = I^{g'}$}
                                     \State $I^{g'} \leftarrow I^g$ 
                                    \For{each $e': a' \rightarrow^{\exists} c' \in S^{\exists}$}
                                        \State $M^{e'} \leftarrow \llbracket \mathds{Q}(a')\rrbracket_{\mathds{S}(I^g)}$
                                        \For{each $m' \in M^{e'}$}
                                            \If{$\llbracket \mathds{Q}(m'(c'))\rrbracket_{\mathds{S}(I^g)} \not = \emptyset$}
                                                \State $g^e \leftarrow$ mapping from $vars(m'(c))$ to fresh \texttt{IRI}s
                                                \State $I^g \leftarrow I^g \cup g^e(m'(c))$
                                            \EndIf
                                        \EndFor
                                    \EndFor
                                    \EndWhile
                                    \State $I^g \leftarrow clos(I^g,R)$
                                     \State $M^{I} \leftarrow \llbracket \mathds{Q}(a)\rrbracket_{\mathds{S}(I^g)}$
                                    \For{each $m^{I} \in M^{I}$}
                                        \If{$\llbracket \mathds{Q}(m^{I}(c))\rrbracket_{\mathds{S}(I^g)} \not = \emptyset$}
                                            \State  $V \leftarrow V \cup \{e\}$ 
                                        \EndIf
                                    \EndFor
                            \EndFor
                        \EndFor
                    \EndIf
                \EndFor
        \EndFor
        \Return $S^{\protect\exists} \setminus V$
        \EndProcedure
    \end{algorithmic}
\end{algorithm}

We can achieve finding these violating instances if they exist,  intituitively, by starting from triples that are (1) groundings of the inference rules' antecedents, (2) instances of the original schema $S$, and (3) which produce, via the closure, a fact on which we can map $a$. To find the minimal number of inference rules' antecedents that we have to ground, we can reason ``backwards'' starting from an inference rule antecedent $A$ whose consequent can trigger $e$, and compute inference rules' antecedents than can compute $A$. We have implemented this backward-chaining reasoning in a way similar to query rewriting in OBDA~\cite{Calvanese2009}, and the Query-Sub-Query algorithm in datalog~\cite{abiteboulbook}. We don't provide the specifics of the algorithm but emphasize that it terminates by relying on a notion of minimality of the rewritings produced. A rewriting produced by our algorithm is  essentially a ``transitive'' antecedent via our inference rules, which can produce $A$. By instantiating these rule antecedents in one rewriting, that is also an instance of $\langle S^{G},S^{\Delta},\emptyset \rangle$, and ``closing'' it existentially\footnote{For this step we implement a version of the chase algorithm~\cite{benedikt2017benchmarking}.} with $S^{\exists}$ we produce such a ``minimal'' instance of the original schema on the closure of which we know we can find $A$. This $A$ is the antecedent of an inference rule that can infer facts matching the antecedent of $e$, and thus, after applying this rule, we can check the satisfaction or violation of $e$. Our rewritings' groundings are representative of all possible instances whose closure can lead to producing a fact that the antecedent of $e$ maps to; if $e$ is valid in all these instances then $e$ can not be violated in any closure of an instance of $S$, and thus we retain it from $S^{\exists}$.

The pseudocode for our algorithm can be seen in Algorithm \ref{algExistential}. For each existential rule $e$ we consider each inference rule $r: A \rightarrow C$ such that inferring $C$ could trigger $e$ (lines 3-5). We then compute all the rewritings $W$ of $A$ by means of backward-chaining the rules in $R$. For each such rewriting $A^w$, we want to see if we can match it on the instances of $S$. We do so reusing the \score\ approach, by computing the set of mappings $M^w = \llbracket \mathds{Q}(A^w)\rrbracket_{\mathds{S}(S)}$. If $M^w$ is empty, then there is no instance of $S$ on which $A^w$ would match. Otherwise, we consider whether this potential match could violate the existential rule $e$ (lines 10-26). For each mapping $m^w \in M^w$ we compute the instance $I^g$, grounding of $A^w$, by first applying to $A^w$ all mappings in $m^w$ that do not map a variable to \urilam , and then mapping any remaining variable to fresh \texttt{IRI}s (lines 10-12). To make sure that $I^g$ is an instance of $S$ we perform the chase on $I^g$ using the existential rules. Lines 13 to 21 exactly implement the well-known chase algorithm \cite{benedikt2017benchmarking} to compute existential closure using our own \score\ approach. Finally, we compute the closure on $I^g$ with the inference rules $R$ and, if it violates $e$, we add $e$ to the set $V$ of the existential rules that can be violated (lines 22-26). The output of our algorithm ($S'^{\exists}$) is $S^{\exists} \setminus V$.

\section{Experimental Evaluation}

We developed a Java implementation of our approaches. This allowed us to test their correctness with a number of test cases, and to compare their scalability by evaluating them on synthetic schemas of different sizes. We present here two experiments. In the first one, we compare the computational time to compute the simple schema consequence, on schemas of different size, using the \score\ and \critical\ approaches. In the second one, we show the overhead in computational time to compute the existential-preserving schema consequence. Since this overhead is the same, regardless of which approach we use to compute the simple schema consequence, we chose to use \score.

We developed a synthetic schema and an inference rule generator that is configurable with 8 parameters: $\pi_C,\lvert P\rvert,\lvert U\rvert,\lvert L\rvert,\lvert S^G\rvert,\lvert R\rvert, \lvert S^{\exists}\rvert, n_A$, 
which we now describe. To reflect the fact that triple predicates are typically defined in vocabularies, our generator does not consider variables in the predicate position. Random triple patterns are created as follows. Predicate \texttt{IRI}s are randomly selected from a set of \texttt{IRI}s $P$. Elements in the subject and object position are instantiated as constants with probability $\pi_C$, or else as new variables. Constants in the subject positions are instantiated with a random \texttt{IRI}, and constants in the object position with a random \texttt{IRI} with $50\%$ probability, or otherwise with a random literal. Random \texttt{IRI}s and literals are selected, respectively, from sets $U$ and $L$ ($U \cap P = \emptyset$). We consider chain rules where the triples in the antecedent join each other to form a list where the object of a triple is the same as the subject of the next. The consequent of each rule is a triple having the subject of the first triple in the antecedent as a subject, and the object of the last triple as object. An example of such inference rule generated by our experiment is: 
$\{ \trip{\var{v0}}{\uri{m1}}{\var{v1}}, \trip{\var{v1}}{\uri{m3}}{\var{v2}} \} \rightarrow \{\trip{\var{v0}}{\uri{m2}}{\var{v2}}\} $
In each run of the experiment we populate a schema $S = \langle S^G, S^{\Delta}, S^{\exists}\rangle$ and a set of inference rules $R$ having $n_A$ triples in the antecedent. To ensure that some inference rules in each set are applicable, half of the schema is initialized with the antecedents triples of randomly selected inference rules. The other half is populated with random triple patterns. Each existential rule of schema $S$ is created as follows. Its antecedent is selected randomly from all the consequents of the inference rules, while its consequence is selected randomly from all the antecedents of all the inference rules. This is done to ensure the relevance of the existential rules, and increase the likelihood of interactions with the inference rules. We initialize $S^\Delta$ with all the variables in the subject and predicate position in the triples of $S$. The code for these experiments is available on GitHub.\textsuperscript{\ref{externalImplementationLink}} 
We run the experiments on a standard Java virtual machine running on Ubuntu 16.04 with 15.5 GB RAM, an Intel Core i7-6700 Processor. Average completion times of over 10 minutes have not been recorded.

\begin{figure}[t]

        \centering{\subfloat[]{%
        \includegraphics[clip,width=0.46\columnwidth]{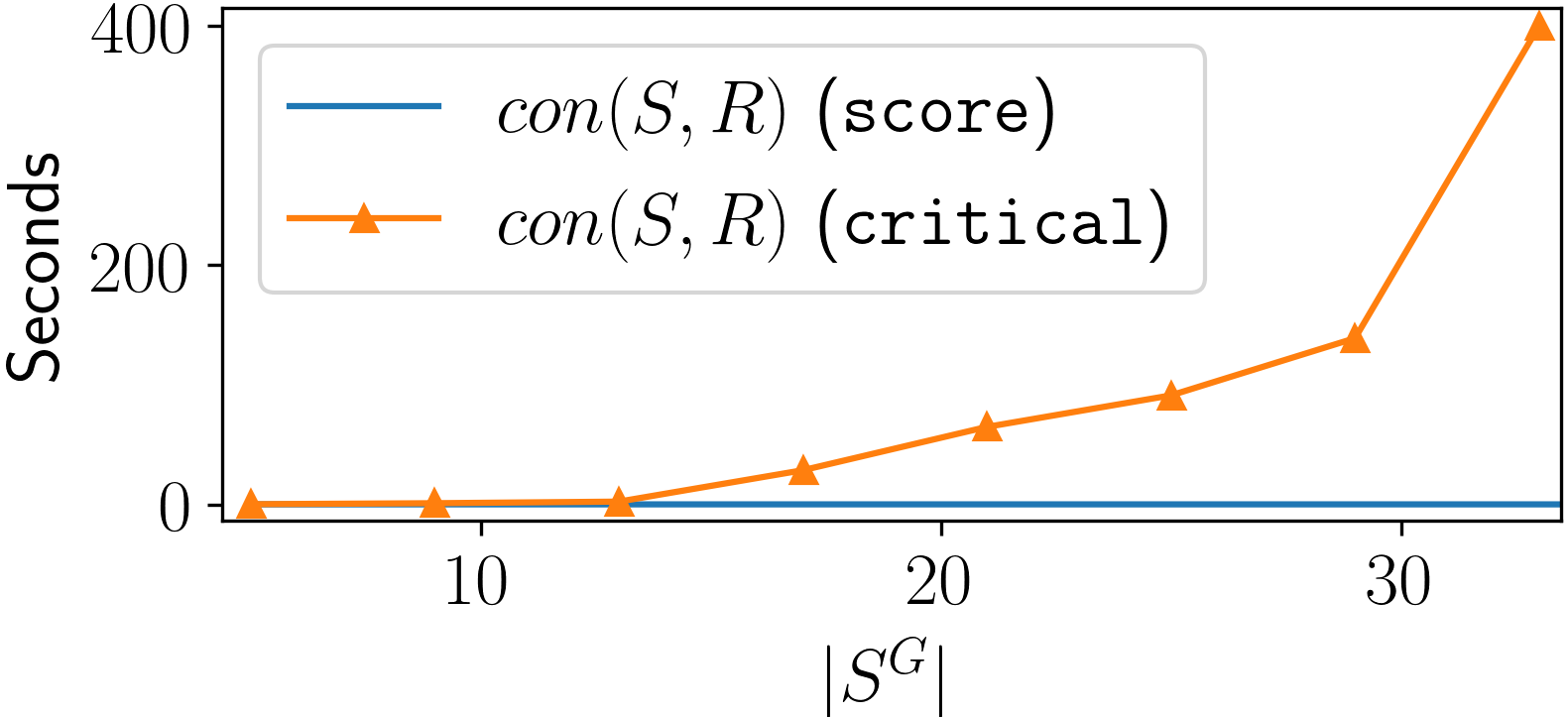}%
        }}
        \qquad
        \subfloat[]{%
        
        \includegraphics[clip,width=0.46\columnwidth]{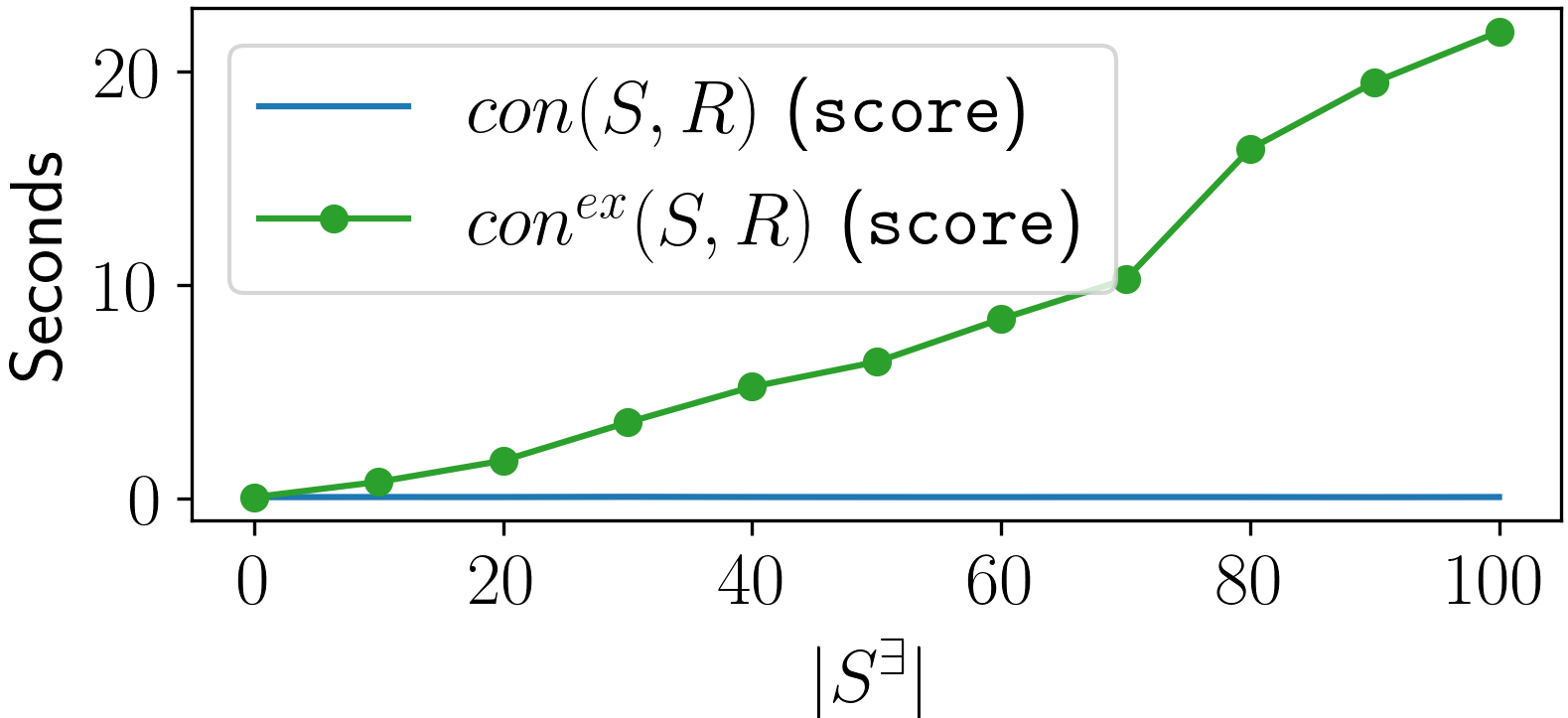}%
        
        }
        \caption{\label{fig:my-label} (a) Average time to compute $con(S,R)$ using \score\ and \critical\ as the schema size $\lvert S^G\rvert$ grows. 
        The other parameters are: $\lvert P\rvert=1.5\lvert S^G\rvert$, $\pi_C=0.1$, $\lvert U\rvert =\lvert L\rvert =\lvert S^G\rvert$, $\lvert R\rvert =4$, $n_A=2$, $\rvert S^{\protect\exists{}}  \lvert =0$. (b) Average time to compute $con(S,R)$ and $con^{ex}(S,R)$ using the \score\ approach as the number of existential rules $\rvert S^{\protect\exists} \lvert $ increases. The other parameters are $\lvert S\rvert = 100$, $\lvert P\rvert= 110$, $\pi_C=0.1$, $\lvert U\rvert =\lvert L\rvert =\lvert S^G\rvert$, $\lvert R\rvert =20$, $n_A=2$.}
\end{figure}

The results of the first experiment are displayed in Fig.\ \ref{fig:my-label} (a). This figure shows the time to compute the schema consequence for different schema sizes $\lvert S\rvert$ using \score\ and \critical . The parameters have been chosen to be small enough to accommodate for the high computational complexity of the \critical\ approach.  This figure shows that \score\ is orders of magnitude faster, especially on large schema sizes. The \critical\ approach, instead, times out for schema sizes of over 33 triples. 
Fig.\ \ref{fig:my-label} (b) shows the increase of computation time as schemas with more existential rules are considered. The results of the second experiment show how our approach to compute the existential-preserving schema consequence can scale to a large number of existential rules on a large input schema in a matter of seconds. We should note that, given the complexity of \texttt{SPARQL} query answering \cite{Perez2006Semantics}, we can expect an exponential increase in computation time as more triples are added to the antecedent of a rule.

\section{Conclusion}
SHACL constraints can can be used to define the schema of graph datasets. However, the application of inference rules could cause the violation of such constraints, and thus require a change in the schema. In this paper we address the problem of computing the \emph{schema consequence} of a schema $S$ and a set of rules $R$, that is, the evolved schema of the graphs, instances of $S$, closed under inference rules $R$. To address this problem we introduced our notion of a \emph{triplestore schema}, which captures a fragment of SHACL, and can also be used as a standalone logical tool to model properties of RDF graphs in general. We addressed the problem incrementally, first by computing a \emph{simple} schema consequence that does not consider existential constraints. We presented two approaches to compute the simple schema consequence. The first is based on the pre-existing concept of a \emph{critical instance}, while the second is a novel approach based on query rewriting and which our experiments showed to be significantly more efficient. We have then provided an approach to deal with existential constraints based on backward-chaining reasoning, which computes what we call an \emph{existential-preserving} schema consequence. This can be considered the final output of our approach, which a domain expert can use to update the schema of an \texttt{RDF} dataset, if they choose to retain all the inference rules considered. The machinery we developed in the form of the simple schema consequence, can also have other applications, such as determining which rules are applicable on a dataset and, if they are, what kind of triples they can infer.

\section*{Acknowledgments}
This work was supported by an Institutional Links grant, ID 333778, under the Newton-Katip Çelebi Fund. The grant is funded by the UK  Department for Business, Energy and Industrial Strategy and the Scientific and Technological Research Council of Turkey (TUBITAK) under grant 116E918, and delivered by the British Council.

\bibliographystyle{splncs04}
\bibliography{litbib}

\begin{thebibliography}{10}
\providecommand{\url}[1]{\texttt{#1}}
\providecommand{\urlprefix}{URL }
\providecommand{\doi}[1]{https://doi.org/#1}

\bibitem{abiteboulbook}
Abiteboul, S., Hull, R., Vianu, V.: Foundations of databases: the logical
  level. Addison-Wesley Longman Publishing Co., Inc. (1995)

\bibitem{baget2011rules}
Baget, J.F., Lecl{\`e}re, M., Mugnier, M.L., Salvat, E.: On rules with
  existential variables: Walking the decidability line. Artificial Intelligence
   \textbf{175}(9-10),  1620--1654 (2011)

\bibitem{Nick2018SWRL2SPIN}
Bassiliades, N.: {SWRL2SPIN:} {A} tool for transforming {SWRL} rule bases in
  {OWL} ontologies to object-oriented {SPIN} rules. CoRR  (2018),
  \url{http://arxiv.org/abs/1801.09061}

\bibitem{benedikt2017benchmarking}
Benedikt, M., Konstantinidis, G., Mecca, G., Motik, B., Papotti, P., Santoro,
  D., Tsamoura, E.: Benchmarking the chase. In: Proceedings of the 36th ACM
  SIGMOD-SIGACT-SIGAI Symposium on Principles of Database Systems. pp. 37--52.
  ACM (2017)

\bibitem{Calvanese2009}
Calvanese, D., De~Giacomo, G., Lembo, D., Lenzerini, M., Poggi, A.,
  Rodriguez-Muro, M., Rosati, R.: Ontologies and Databases: The DL-Lite
  Approach, pp. 255--356. Springer Berlin Heidelberg, Berlin, Heidelberg (2009)

\bibitem{Ceri1989Datalog}
Ceri, S., Gottlob, G., Tanca, L.: What you always wanted to know about datalog
  (and never dared to ask). IEEE Transactions on Knowledge and Data Engineering
   \textbf{1}(1),  146--166 (1989)

\bibitem{Corman2018SHACL}
Corman, J., Reutter, J.L., Savkovi{\'{c}}, O.: {Semantics and Validation of
  Recursive SHACL}. In: The Semantic Web -- ISWC 2018. pp. 318--336 (2018)

\bibitem{fagin2005data}
Fagin, R., Kolaitis, P.G., Miller, R.J., Popa, L.: Data exchange: semantics and
  query answering. Theoretical Computer Science  \textbf{336}(1),  89--124
  (2005)

\bibitem{glimm2014abstraction}
Glimm, B., Kazakov, Y., Liebig, T., Tran, T.K., Vialard, V.: Abstraction
  refinement for ontology materialization. In: International Semantic Web
  Conference. pp. 180--195 (2014)

\bibitem{2017SHACL}
Knublauch, H., Kontokostas, D.: {Shapes constraint language (SHACL)}. {W3C}
  {R}ecommendation, {W3C} (2017), \url{https://www.w3.org/TR/shacl/}

\bibitem{Taylor2017SSN}
Lefran\c{c}ois, M., Cox, S., Taylor, K., Haller, A., Janowicz, K., Phuoc, D.L.:
  {Semantic Sensor Network Ontology}. {W3C} {R}ecommendation, W3C (2017),
  \url{https://www.w3.org/TR/2017/REC-vocab-ssn-20171019/}

\bibitem{marnette2009generalized}
Marnette, B.: Generalized schema-mappings: from termination to tractability.
  In: Proc. of the twenty-eighth ACM SIGMOD-SIGACT-SIGART symp. on Principles
  of database systems. pp. 13--22. ACM (2009)

\bibitem{Perez2006Semantics}
P{\'e}rez, J., Arenas, M., Gutierrez, C.: {Semantics and Complexity of SPARQL}.
  ACM Transactions on Database Systems  \textbf{34}(3),  16:1--16:45 (2009)

\bibitem{Prudhommeaux2014ShapeExpressions}
Prud'hommeaux, E., Labra~Gayo, J.E., Solbrig, H.: {Shape Expressions: An RDF
  Validation and Transformation Language}. In: Proceedings of the 10th
  International Conference on Semantic Systems. pp. 32--40. SEM '14, ACM (2014)

\end{thebibliography}

\appendix

\section{Proof of Theorem \ref{theorem:approachEquality}}

\begin{customthm}{1} \label{theorem:approachEquality}
  For all rules $r: A \rightarrow C$ and triplestore schemas $S$, $\mathds{I}(\score(S,r)) = \\ \mathds{I}(\critical(S,r))$.
\end{customthm}

\begin{Proof} Since the mappings generated by the \critical\ and \score\ approaches (namely $\llbracket A\rrbracket_{\mathds{C}(S,r)}$ and $\llbracket \mathds{Q}(A)\rrbracket_{\mathds{S}(S)}$) are post-processed in the same way, we demonstrate the equivalence of the two approaches by demonstrating that ($\Leftarrow$) the set of mappings generated by $\score(S,r)$, which are not filtered out by our post-processing of the mappings, is a subset of the mappings generated by $\critical(S,r)$ and that ($\Rightarrow$) every mapping $\critical(S,r) \setminus \score(S,r)$ is redundant with respect to another mapping in $\critical(S,r) \cap \score(S,r)$. We denote with $\Omega(m,A,S)$ the set of mappings that are not filtered out by our filtering function with respect to $A$ and $S$. We say that a mapping $m$ is redundant with respect of a set of mappings $M$ (and a schema $S$ and rule $r$) if the schema consequences computed over $M$ and $' \setminus \{m\}$ are semantically equivalent.

Note that both the critical and the sandbox instance are constructed by taking a triple in schema $S^G$ and substituting its variables. Thus we can associate each triple in the critical or the sandbox instance with the triples in $S$ that can be used to construct them, and we will call the latter the \emph{origin} triples of the former. For each triple in the critical or the sandbox instance, at least one origin triple must exist. 

$\Rightarrow$) Let a mapping $m \in \llbracket A\rrbracket_{\mathds{C}(S,r)} \setminus \llbracket \mathds{Q}(A)\rrbracket_{\mathds{S}(S)}$. We get all triples $t_A$ of $A$ and for each one we will construct a triple $t_q$ of a conjunctive query $q \in \mathds{Q}(A)$, and mapping $m^{*}$ of $q$ into $\mathds{S}(S)$, such that either $m = m^{*}$ or $m^{*}$ makes $m$ redundant. 
For $i \in \tau$, if $t_A[i]$ is a variable $?v$ and $m(?v)$ $=$ $\urilam$ then set $t_q[i] = ?v$, since there must be a triple in $\mathds{S}(S)$ with $\urilam$ in the $i^{th}$ position so if $m^{*}$ mapped $t_q$ on that triple, $m^{*}(t_q[i]) = m(t_A[i]) = \urilam$ (for all triples of the critical instance that have $\urilam$ in position $i$, its origin triples have variables in the same position, so the sandbox instance would also have $\urilam$ in position $i$). If $t_A[i]$ is a variable $?v$ and $m(?v)$ is a constant $c$, then we distinguish two cases (a) and (b). 
Let $t_S \in S$ be an origin triple of $m(t_A)$ in the critical instance, then (a) if $t_S[i]=c$ then $\mathds{S}(S)$ would have retained $c$ in the corresponding position in the triple of which $t_S$ is the origin, and so we set $t_q[i]$ to $?v$, in order for $(?v$ $\rightarrow$ $c)$ $\in$ $m$ to also belong to mapping $m^{*}$ of $t_q$ into $\mathds{S}(S)$, or 
(b) if $t_S[i]$ is a variable (we know that $\urilam$ is the element in position $i$ of the triple in the sandbox graph of which $t_S$ is an origin) we consider two sub-cases ($b^{\prime}$) and ($\neg b^{\prime}$). 

We set $t_q[i]$ to $\urilam$ in case ($b^{\prime}$), namely if there is a position $j$ in a triple $t_A^{\prime} \in A$ (different position to $i$, i.e., $j\neq i$ if $t_A = t_A^{\prime}$), for which $t_A^{\prime}[j] = t_A[i]$ and for $t_S^{\prime}$, an origin triple of $t_A^{\prime}$, $t_S^{\prime}[j] = c$. Condition ($b^{\prime}$) essentially says that $?v$ will have to be mapped to $c$ in the sandbox graph due to some other position of the rewriting, so $t_q[i]$ can just be set to $\urilam$ to simply match the corresponding triple in $\mathds{S}(S)$. In case ($\neg b^{\prime}$) we set $t_q[i]$ to $?v$; this condition produces a more \emph{general} mapping $m^{*}$ since $m(?v) = c$ while $m^{*}(?v)$ will be $\urilam$. We say that $m_2$ is more general than $m_1$, denoted by $m_1 \leq_{\lambda} m_2$, if $dom(m_1)$ $=$ $dom(m_2)$ and for all $?v \in dom(m_1)$, either $m_2(?v) = m_1(?v)$ or $m_2(?v)$ $=$ $\urilam$.

Lastly we consider the case of $t_A[i]$ being a constant. If there is an origin triple $t_S$ of $m(t_A)$ in the critical instance such that $t_S[i]$ is the same constant  we also set it to $t_q[i]$. Otherwise, if $t_S[i]$ is a variable in any origin triple $t_S$ we set $t_q[i]$ to $\urilam$. This does not change the mapping under construction $m^{*}$. By following the described process for all triple in $A$ we end up with a rewriting $q$ of $A$ and a mapping $m^{*}$ from $q$ to $ \mathds{S}(S)$ such that $m^{*}$ is more general than $m$.

The existence of the more general mapping $m^{*}$ makes mapping $m$ redundant. This is true because if $m$ is not filtered out by our post-processing, then $m^{*}$ would also not be filtered out (since mappings can only be filtered out only when a variable is mapped to a literal, but $m^{*}$ does not contain any such mapping not already in $m$). If $m$ is not filtered out, then this would lead to the schema being expanded with triples $m(C)$. The instances of the schema obtained through mapping $m$ are a subset of those obtained through mapping $m^{*}$. This can be demonstrated by noticing that for every triple in $m(C)$ there is a corresponding triple in $m^{*}(C)$ that has at each position, either the same constant $c$ or $\urilam$, which is then substituted with a variable in the schema consequence. This variable can always be substituted with the same constant $c$, when instantiating the schema, and therefore there cannot be an instance of the schema consequence generated from mapping $m$ that is not an instance of the schema consequence generated from mapping $m^{*}$. In fact, the only case in which a variable $?v$ in a schema cannot be instantiated with a constant $c$ is when $c$ is a literal and $?v$ is in the no-literal set. However, for all mappings $m \in \llbracket A\rrbracket_{\mathds{C}(S,r)} \setminus \llbracket \mathds{Q}(A)\rrbracket_{\mathds{S}(S)}$, we can notice that whenever our post-processing function adds a variable to the no-literal set, it would have rejected the mapping if that variable was mapped to a literal instead.

$\Leftarrow$) Let a mapping $m^{*} \in \llbracket \mathds{Q}(A)\rrbracket_{\mathds{S}(S)}$. This means that there is a $q \in \mathds{Q}(A)$, for which $m^{*} \in \llbracket q\rrbracket_{\mathds{S}(S)}$. We get all triples $t_q$ of $q$ and for each one we will show that for the triple $t_A$ in $A$, that was rewritten into $t_q$, it holds that either $m^{*}$ is filtered out by the filtering function $\Omega$, or it is also a mapping from $t_A$ into the critical instance (i.e.\ $m^*(t_A) \in \mathds{C}(S,r)$), hence $m^{*} \in \llbracket A\rrbracket_{\mathds{C}(S,r)}$. 
This would prove that the mappings generated by $\llbracket \mathds{Q}(A)\rrbracket_{\mathds{S}(S)}$ are either subsumed by the mappings in $\llbracket A\rrbracket_{\mathds{C}(S,r)}$, or they would be filtered out anyway, and thus $\Omega(\llbracket \mathds{Q}(A)\rrbracket_{\mathds{S}(S)},A,S) \subseteq \Omega(\llbracket A\rrbracket_{\mathds{C}(S,r)},A,S)$. 

For all $t_q \in q$, for all  $i \in \tau$, if $t_q[i]$ is a variable $?v$ then this position has not been changed from rewriting $t_A$ into $t_q$, thus $?v$ will exist in $t_A[i]$; 
if $m^{*}(?v)$ is $\urilam$, then $m^{*}(t_q)[i] = \urilam$, and therefore in any origin triple $t_S$ of $m^{*}(t_q)$, $t_S[i]$ had to be a variable and so $\urilam$ will exist in position $i$ in the triples the critical instance will create for $t_S$, and for these triples $m^{*}$ will partially map $t_A$ onto them. 
Similarly, if $t_q[i]$ is a variable $?v$ and $m^{*}(?v)$ is a constant $c$, then this constant would be present in $t_S[i]$ and so also in all triples coming from $t_S$ in the critical instance; again $m^{*}$ will partially map $t_A$ onto the triples in the critical instance generated from $t_S$. 

If $t_q[i]$ is a constant other than \urilam , then the triple $t_A$ that got rewritten to triple $t_q$ would have the same constant in position $i$. Also, this constant must be present in position $i$ in the triple $m^{*}(t_q)$ of the sandbox graph, and therefore in position $i$ of any of its origin triples $t_S$. In fact, by virtue of how the sandbox graph is constructed, any triple of the schema that does not have this constant in position $i$, cannot be an origin triple of $m^{*}(t_q)$. 
Thus again, $m^{*}$ will partially map $t_A$ onto the triples in the critical instance of which $t_S$ is the origin triple. 

Lastly, if $t_q[i]$ is $\urilam$, then  $m^{*}(t_q)[i] = \urilam$, and in any origin triple $t_S$ of the sandbox triple $m^{*}(t_q)$ we have a variable in position $i$, and in the critical instance we have triples, of which $t_S$ is the origin, with all possible \texttt{URI}s or $\urilam$ in this position; this means that if $m^*(t_A[i])$ is a \texttt{URI}, in the triple $t_A$ that got rewritten in $t_q$, we can ``match'' it in the critical instance: 
if $t_A[i]$ is a variable $?v_1$ then $t_A$ will match the triple in the critical instance (instance of any origin triple $t_S$ of $m^{*}(t_q)$) that contains $m^{*}(?v_1)$ in position $i$ (notice that $m^{*}$ should contain a value for $v_1$ since all variables are present in all rewritings of $A$); else, if $t_A[i]$ is a \texttt{URI} there will be a corresponding image triple (again instance of any $t_S$)  in the critical instance with the same constant in this position. 

If, however, $m^*(t_A[i])$ is a literal, it is possible that there is no triple in the critical instance that can match the literal $m^*(t_A[i])$ in position $i$. However we will show that, in this case, the filter function $\Omega$ will remove this mapping. If literal $m^*(t_A[i])$ does not occur in position $i$ in any of the triples in the critical instance that have the same origin triple as $m^*(t_q)$ then, by the definition of how the critical instance is constructed, this must be caused by the fact that either (a) $i \in \{1,2\}$ or (b) for all the possible origin triples $t_S$ of $m^*(t_q)$, $t_S[i] \in S^\Delta$ or $t_S[i]$ is a constant other than $m^*(t_A[i])$. In both cases, our filtering function will filter out the mapping. In fact, it filters out mappings where variables are mapped to literals, if they occur in the subject or predicate position of a triple in $A$. This is true in case (a). Moreover it filters out mappings if a variable in the object position there is a literal $l$, or a variable mapped to a literal $l$, such that there is no origin triple $t_S$ of $m^*(t_q)$ such that $t_S[i] = l$ or $t_S[i]$ is a variable not in $S^{\Delta}$. This is true in case (b). Thus $m^{*}$ is a mapping from $t_A$ into the critical instance. \end{Proof}

\section{Proof of Theorem \ref{newTheoremSimple}}

\begin{customthm}{2} \label{newTheoremSimple} For all rules $r: A \rightarrow C$ and triplestore schemas $S$, $\mathds{I}(\score(S,r)) = \\ \mathds{I}(r(S))$.

\end{customthm}

Theorem \ref{newTheoremSimple} states that $\score(S,r)$ is semantically equivalent to $r(S)$. We prove this by showing that ($\Rightarrow$) every triple that can be inferred by applying rule $r$ on an instance of $S$ is an instance of the schema $\score(S,r) \setminus S$ and that ($\Leftarrow$) every triple in every instance of schema $\score(S,r) \setminus S$ can be obtained by applying rule $r$ on an instance of $S$. To reduce the complexity of the proof, we consider the case without literals (i.e. when no literal occurs in $S$ and $r$, and all of the variables in the schema $S^G$ are in $S^{\Delta}$). This proof can be extended to include literals by considering the the post-processing of the mappings, as demonstrated in the proof of Theorem \ref{theorem:approachEquality}. 
More precisely, we reformulate Theorem \ref{newTheoremSimple} as follows:

\begin{lem} \label{lemmaOldTheorem} For all rules $r: A \rightarrow C$ and for all triplestore schemas $S$ such that $S^\Delta = vars(S^G)$ and that no literal occurs in $r$ and $S^G$:

$\Rightarrow$) for all triple patterns $\Nto \in r(S) \setminus S$, 
for all triples $\Ntoc$ in an instance of \Nto, 
there exists a triple pattern $\NtoPrime \in \score(S,r) \setminus S$ 
s.t. $\{\Ntoc\} \in \mathds{I}(\{\NtoPrime\})$

$\Leftarrow$) for all triple patterns $\NtoPrime \in \score(S,r) \setminus S$, for all triples $\Ntoc$ in an instance of  $\NtoPrime$ there exists a triple pattern $\Nto \in r(S) \setminus S$ s.t. $\{\Ntoc\} \in \mathds{I}(\{\Nto\})$.

\end{lem}

\begin{Proof}
$\Rightarrow$) Given the premises of this part of Lemma \ref{lemmaOldTheorem}, there must exist a graph $I \in \mathds{I}(\{S\}$ on which an application of rule $r$ generates $\Ntoc$. For this to happen, there must exist a mapping $m \in \llbracket A\rrbracket_{I}$, such that $\Ntoc \in m(C)$. Obviously, the set of triples on which $A$ matches into $I$ via $m$ is $m(A) \subseteq I$.

For all triples $\Nta \in m(A)$ there exists a triple pattern $\Ntao \in S^G$ that models $\Nta$ (e.g.\ $\{\Nta\} \in \mathds{I}(\Ntao)$ ). We choose one of such triple pattern $\Ntao$ (in case multiple ones exist) and we call it the \emph{modelling triple} of $\Nta$. To prove the first part of Lemma \ref{lemmaOldTheorem} we will make use of the following proposition:

\emph{There exists a mapping $m^*$ and a rewriting $q \in \mathds{Q}(A)$, such that $m^*(q) \subseteq \mathds{S}(S)$ (thus $m^* \in \llbracket q\rrbracket_{\mathds{S}(S)}$), and that $m$ and $m^*$ agree on all variables except for those that are mapped to $\urilam$ in $m^*$. Formally, $dom(m) = dom(m^*)$ and for every variable $?v$ in $dom(m^*)$, either $m^*(?v) = m(?v)$, or $m^*(?v) = \urilam$.}

By proving the proposition above we know that our algorithm would extend the original schema graph $S^G$ with graph pattern $unpack(m^*(C))$, and the set of variables $S^\Delta$ with $vars(unpack(m^*(C)))$, where $unpack$ is a function that substitutes each $:\lambda$ in a graph with a fresh variable. If we take a triple pattern $c \in C$ such that $\Ntoc = m(\{c\})$, then $unpack(m^*(\{c\}))$ belongs to $\score(S,r)^G \setminus S^G$ (because $m^*$ matches a rewriting of the antecedent of the rule $r$ on the sandbox graph). Graph $unpack(m^*(\{c\}))$ also models triple $\Ntoc$. This is true because for each of the three positions $i$ of the schema triple pattern $c$, either $c[i]$ is a constant, and therefore this constant will also appear in position $i$ in both $unpack(m^*(\{c\}))[i]$ and $\Ntoc[i]$ or it is a variable, and by our proposition above either (a) $m^*(c)[i] = m(c)[i] = \Ntoc[i]$ and so $unpack(m^*(\{c\}))[i] = \Ntoc[i]$, or (b) $m^*(c)[i] = \urilam$ and therefore $unpack(m^*(\{c\}))[i]$ is a variable. We can then trivially recreate $\Ntoc$ as an instance of $unpack(m^*(\{c\}))$ by assigning to each variable $?v$ in $unpack(m^*(\{c\}))[i]$ the value $\Ntoc[i]$. Thus our algorithm would generate a triple pattern that models $\Ntoc$ and that would complete the proof of the first part ($\Rightarrow$) of the Theorem.

The proof of our proposition is as follows. Consider every position $i$ in every triple $\Nta \in m(A)$, and their corresponding modelling triple $\Ntao \in S$. Let $\NtA$ be the triple in $A$ such that $m(\NtA) = \Nta$. Let $\Nts$ be $\mathds{S}(\Ntao)$, and thus $\Nts \in \mathds{S}(S)$. We are now going to construct a query $q$, rewriting of $\mathds{Q}(A)$, such that $\llbracket q\rrbracket_{\mathds{S}(S)}$ gives us the mapping $m^*$ in our proposition. 
And to do this, we have to construct a triple pattern $\Ntq$ which will be the rewriting of $\NtA$ in $q$. By definition of how our rewritings are constructed, every element $\Ntq[i]$ is either $\NtA[i]$ or $\urilam$.

To set the value of each element in the triple pattern $\Ntq$ we consider the four possible cases that arise from $\NtA[i]$ and $\Ntao[i]$ being either a constant or a variable. In parallel, we consider the element-by-element evaluation of $q$ on $\mathds{S}(S)$ which generates $m^*$. 
\begin{enumerate}
    \item If $\NtA[i]$ and $\Ntao[i]$ are both constants, then since $\Nta \in m(A)$, it must be true that $\NtA[i] = \Nta[i]$. Moreover, since $\Ntao$ must be a model of $\Nta$, it follows that $\Ntao[i] = \Nta[i]$ and therefore $\Nts[i] = \Nta[i]$. We set element $\Ntq[i]$ to be the constant $\NtA[i]$, which matches the constant $\Nts[i]$.
    \item If $\NtA[i]$ is a constant but $\Ntao[i]$ is a variable, then we know that $\Nts[i] = \urilam$. We set element $\Ntq[i]$ to be the constant $\urilam$, which matches the constant $\Nts[i]$.
    \item If $\NtA[i]$ is a variable $?v$ and $\Ntao[i]$ is a constant $x$, then we know that $\Nta[i] = \Nts[i] = x$. Therefore mapping $m$ must contain binding $?v \to x$. We set element $\Ntq[i]$ to be the variable $?v$, so that when we complete the construction of $q$, $m^*$ will contain mapping $?v \to x$.
    \item If $\NtA[i]$ and $\Ntao[i]$ are both variables, it must be that $\Nts[i] = \urilam$.
    If it exists a triple $\NtAPrime \in A$ and a modelling triple $\NtAoPrime$ of $m(\NtAPrime)$, and position $j$ such that $\NtAPrime[i]$ is variable $?v$ and $\NtAoPrime[i]$ is a constant, then we set element $\Ntq[i]$ to be the constant $\urilam$ ($\Nts[i]$ will be $\urilam$ in our sandbox graph). Note that even though we don't use variable $\NtAPrime[i]$ in this part of our rewriting, the aforementioned existence of triple $\NtAPrime$ will force $m^*$ to contain a binding for this variable, and this binding will be in $m$ because this pair $\NtAPrime[i]$ and  $\NtAoPrime[i]$ will be treated by case $3$.
     Otherwise we set element $\Ntq[i]$ to be the variable $\NtA[i]$, which will bind with the value $\urilam$ of $\Nts[i]$ generating binding $?v \to \urilam$. 
\end{enumerate}
This proves our proposition, as we have constructed a query $q$, which belongs to $\mathds{Q}(A)$, which matches $\mathds{S}(S)$ and generates a mapping $m*$ which agrees on $m$ on all variables, except for those which $m^*$ maps to $\urilam$. 

$\Leftarrow$) 
Given the premises of the second direction of Lemma \ref{lemmaOldTheorem}, there must exist a mapping $m^*$ and a query $q \in \mathds{Q}(A)$ such that: $\NtoPrime \in unpack(m^*(C))$, $m^*(q) \subseteq \mathds{S}(S)$ and $dom(m^*) = dom(A)$. Note that $\mathds{S}(S)$ is an instance of $S$. There must exist a triple pattern $\Ntc$ in $C$ such that $unpack(m^*(\{\Ntc\})) = \NtoPrime$.

Note that $unpack$ transforms a triple  in a triple pattern by changing occurrences of $\lambda$ into fresh variables, if no $\lambda$ occurs in the triple, the triple remains unchanged.

For all positions $i$ of $\Ntc$, if $\Ntc[i]$ is a variable $?v$, given that $unpack(m^*(\{\Ntc\}))$ models $\Ntoc$, then mapping $m^*$ either contains the binding $?v \to \Ntoc[i]$, or it contains $?v \to \urilam$. Now consider the mapping $m$ generated by modifying mapping $m^*$ in the following way. For each position $i$ of $\Ntc$, if $\Ntc[i]$ is a variable $?v$ and $?v \to \urilam \in m^*$, substitute this binding with $?v \to \Ntoc[i]$. It is trivial to see that $\Ntoc \in m(C)$. We will now show that this mapping $m$ can be computed by evaluating the original query $A$ on a specific instance of $S$ by showing that $m(A)$ is an instance of $S$. By virtue of how query $q$ is constructed, we know that all of its elements in all of its triples must be the same as $A$, or the constant $\urilam$ instead.

As an intermediate step, we show that $m(q)$ is an instance of $S$. Since we know that $m^*(q)$ is an instance of $S$, and that $\urilam$ cannot occur in the triples in $S$, every occurrence of $\urilam$ in $m^*(q)$ must be generated from a variable $?v$ in a triple in $S$. Since each variable in $S^G$ occurs only once, we can change any occurrence of $\urilam$ in $m^*(q)$ into any other constant and still have an instance of $S$, since, intuitively, in the corresponding positions $S$ contains different variables. Given that the difference between $m$ and $m^*$ is that for some variables that $m^*$ binds to $\urilam$, $m$ binds them to a different constant, it is easy to see that the $m(q)$ can be generated from $m^*(q)$ only by changing some occurrences of $\lambda$ in its triples into a different constant. Thus $m(q)$ is an instance of $S$.

Now we are going to show that $m(A)$ is an instance of $S$. For all triples $\NtAtwo \in A$, our query rewriting algorithm has produced a $\Nta \in a$ for which, for each position $i$, we know that $\Nta[i] = \NtAtwo[i]$ or $\Nta[i] = \urilam$. 
Obviously for $m(\Nta) \in m(q)$, the same holds, i.e., for every position $i$, $m(\Nta)[i] = m(\NtAtwo)[i]$ or $m(\Nta)[i] = \urilam$.  
Since $m(\Nta)$ is an instance of $S$, there must exist a mapping $\gamma$ and a triple pattern $\Nts \in S^G$ such that 
$\gamma(\{\Nts\}) = m(\Nta)$. Since $\Nts$ cannot contain $\urilam$, for every position $i$ where $m(\Nta)[i] = \urilam$, $\Nts[i]$ contains a variable $?v_i$ which does not occur anywhere else in $S^G$ (by virtue of variables occurring only once in the schema); we can obtain the mapping from $\Nts$ to $m(\NtAtwo)$ by substituting each such binding $?v_i \to \urilam$ with $?v_i \to m(\NtAtwo)[i]$ in $\gamma$. For all triples $\NtAtwo \in A$, $m(\NtAtwo)$ is an instance of $S$, and therefore $\bigcup_{\NtAtwo \in A} \{m(\NtAtwo)\}$, which equals $m(A)$, is an instance of $S$.
\end{Proof}

\FloatBarrier

\section{Translation of SHACL constraints into Triplestore Schemas}

In this section of the appendix we present an approach to convert a subset of SHACL constraints into triplestore schemas and vice versa. An implementation of our approaches is available in our code repository.\footnote{\externalimplementationlink}

\subsection{Translation from SHACL constraints to Triplestore Schemas}

We present a simple approach to translate SHACL constraints, captured by a set of shapes, into triplestore schemas. 
For purpose of readability, we present it as a set of templates of SHACL shapes and their corresponding translations in Table \ref{tab:transf}. We use constants \uri{a}, \uri{b}, \uri{c}, \uri{p}, \uri{q}, \uri{s}, \uri{k} and \lit{m} as placeholders constants and ``$...$'' to represent a repetition of elements. The set of SHACL terms that we consider is listed in Figure \ref{fig:shacloperators}. This approach does not deal with all possible shapes constructed with with these terms, and extensions of this approach are possible. For example, it does not deal with cardinality constraints with cardinality different from 1, or recursive shapes.

\begin{figure}
\vspace*{-5mm}
\begin{multicols}{3}
\footnotesize \begin{lstlisting}
$\urish{NodeShape}$
$\urish{targetObjectsOf}$
$\urish{targetSubjectsOf}$
$\urish{targetClass}$
$\urish{nodeKind}$
$\urish{IRI}$
$\urish{IRIOrLiteral}$
$\urish{in}$
$\urish{property}$
$\urish{path}$
$\urish{inversePath}$
$\urish{minCount}$
$\urish{class}$
$\urish{hasValue}$
$\urish{node}$
$\urish{or}$
$\urish{not}$
\end{lstlisting} \normalsize
\end{multicols}
\caption{SHACL terms considered in this approach.}
\label{fig:shacloperators}
\vspace*{-4mm}
\end{figure}

The proposed approach to translate a set of SHACL constraints into a triplestore schema consists of three phases. First, we initialise an empty schema $S = \langle \emptyset, \emptyset, \emptyset \rangle$, and an empty set of uninstantiable predicates $U$. Second, then modify $S$ on a shape-by-shape basis by applying the required transformations as specified in Table \ref{tab:transf}. For each SHACL shape $s$, this involves finding a matching template in the table, namely a template that can be transformed into $s$ by a substitution of the placeholder constants, and then applying the transformations corresponding to that template under the same substitution. Lastly, we remove all triples from $S^G$ whose predicate is in $U$, which we would have populated as we apply the transformations.

We consider three transformations. Given a set of existential constraints $E$, the transformation $include(E)$ simply adds constraints $E$ to $S^{\exists}$. Transformations $allow(T^p, T^{\Delta})$ and $restrict(T^p, T^{\Delta})$ are defined for a set of triple patterns $T^p$, where all the triples in $T^p$ have the same predicate $p$, and a no literal set $T^{\Delta}$. 
Transformations $allow(T^p, T^{\Delta})$ and $restrict(T^p, T^{\Delta})$ add $T^p$ and $T^{\Delta}$ to $S^G$ and $S^{\Delta}$, respectively, if $S^G$ does not contain a triple pattern with $p$. If $S^G$ contains a set of triple patterns $P$ with $p$ as predicate, transformation $restrict(T^p, T^{\Delta})$ removes from $S$ patterns $P$ and their corresponding variables $P^{\Delta}$ from the no-literal set; then adds to $S$ the ``schema intersection'' graph $G'^p$ and its no literal set $ G'^{\Delta}$ where $\mathds{I}(\langle G'^p, G'^{\Delta}, \emptyset \rangle) = \mathds{I}({P, P^{\Delta}, \emptyset}) \cap \mathds{I}({T^p, T^{\Delta}, \emptyset})$. If the schema intersection computed by the $restrict$ transformation is empty, then this predicate cannot occur in the instances of $S$ and we add it to a set of uninstantiable predicates $U$. 

For example, let us imagine a schema $S = \langle \{ \gpattern{\var{v1}}{\uri{a}}{\var{v2}}\}  \}, \{\var{v1}, $\var{v2}$\}, S^{\exists} \rangle$, that is, triples with predicate \uri{a} are allowed in the instances of $S$ if they have an \texttt{IRI} as object. Then consider the application of this transformation,   $restrict(\{\gpattern{\uri{b}}{\uri{a}}{\var{v3}}\}, \{\})$ which restricts triples with predicate \uri{a} to have \uri{b} as subject. The result of this transformation is $\langle \{ \gpattern{\uri{b}}{\uri{a}}{\var{v4}}\}  \}, \{?v4\}, S^{\exists} \rangle$, namely the schema of triples with predicates \uri{a} which have \uri{b} as subject and a \texttt{IRI} as an object.

 \lstset{
   language=C,
   basicstyle=\small,
   breaklines=true
   }


  \begin{center}
 \footnotesize
 \centering
 \begin{longtable}{ | l | l | } 
 \hline
 SHACL & Triplestore Schema \\ 
 \hline
 \hline
 
  \begin{lstlisting}
 $\uri{s}\;\urirdf{type}\;\urish{NodeShape}$; 
   $\urish{targetObjectsOf}\;\uri{p}$$\,;$
   $\urish{nodeKind}\;\urish{IRIOrLiteral}$$\,.$
 \end{lstlisting}  &  \begin{lstlisting}
 $allow(${$\var{v1}\;\uri{p}\;\var{v2}$}, $\{\var{v1}\})$ 
 \end{lstlisting}
 \\ 
 \hline
 
  \begin{lstlisting}
 $\uri{s}\;\urirdf{type}\;\urish{NodeShape}$; 
   $\urish{targetObjectsOf}\;\uri{p}$$\,;$
   $\urish{nodeKind}\;\urish{IRI}$$\,.$
 \end{lstlisting}  &  \begin{lstlisting}
 $restrict(${$\var{v1}\;\uri{p}\;\var{v2}$}, $\{\var{v1}, \var{v2}\})$ 
 \end{lstlisting}
 \\ 
 \hline

 \begin{lstlisting}
 $\uri{s}\;\urirdf{type}\;\urish{NodeShape}$; 
   $\urish{targetObjectsOf}\;\uri{p}$$\,;$
   $\urish{in}$ ($\uri{a}\;\lit{m}\;\ldots$ )$\,.$
 \end{lstlisting}  &  \begin{lstlisting}
 $restrict(${$\var{v1}\;\uri{p}\;\uri{a}$ ,
       $\;\var{v2}\;\uri{p}\;\lit{m}$ ,
       $\;\var{v3}\;\uri{p}$$\,\;\ldots$ }, $\{\var{v1}, \var{v2}, \var{v3}\})$ 
 \end{lstlisting}
 \\ 
 \hline

 \begin{lstlisting}
 $\uri{s}\;\urirdf{type}\;\urish{NodeShape}$; 
   $\urish{targetSubjectsOf}\;\uri{p}$$\,;$
   $\urish{in}$ ($\uri{a}\;\uri{b}\;\ldots$ )$\,.$
 \end{lstlisting}  &  \begin{lstlisting}
 $restrict(${$\uri{a}\;\uri{p}\;\var{v1}$ ,
       $\;\uri{b}\;\uri{p}\;\var{v2}$ ,
       $\;\ldots\;\uri{p}\;\var{v3}$  }, $\{\})$ 
 \end{lstlisting}
 \\ 
 \hline


 \begin{lstlisting}
 $\uri{s}\;\urirdf{type}\;\urish{NodeShape}$; 
   $\urish{targetObjectsOf}\;\uri{p}$$\,;$
   $\urish{class}\;\uri{c}$$\,.$
 \end{lstlisting}  &  \begin{lstlisting}
 $restrict(${$\var{v1}\;\uri{p}\;\var{v2}$}, $\{\var{v1}, \var{v2}\})$ 
 $allow(${$\var{v1}\;\urirdf{type}\;\var{2}$}, $\{\var{v1}\})$
 $include($$\var{v1}\;\uri{p}\;\var{v2}\rightarrow^{\exists}\var{v2}\;\urirdf{type}\;\uri{c})$
 \end{lstlisting}
 \\ 
 \hline

 \begin{lstlisting}
 $\uri{s}\;\urirdf{type}\;\urish{NodeShape}$; 
   $\urish{targetSubjectsOf}\;\uri{p}$$\,;$
   $\urish{class}\;\uri{c}$$\,.$
 \end{lstlisting}  &  \begin{lstlisting}
 $allow(${$\var{v1}\;\uri{p}\;\var{v2}$}, $\{\var{v1}\})$
 $allow(${$\var{v1}\;\urirdf{type}\;\var{v2}$}, $\{\var{v1}\})$ 
 $include($$\var{v1}\;\uri{p}\;\var{v2}\rightarrow^{\exists}\var{v1}\;\urirdf{type}\;\uri{c})$
 \end{lstlisting}
 \\ 
 \hline

 \begin{lstlisting}
 $\uri{s}\;\urirdf{type}\;\urish{NodeShape}$$\,;$ 
   $\urish{targetClass}\;\uri{c}$$\,;$
   $\urish{property}$  [
       $\urish{path}\;\uri{p}$$\,;$
       $\urish{minCount}$ 1$\,;$ ]$\,.$
 \end{lstlisting}  &  \begin{lstlisting}
 $allow(${$\var{v1}\;\urirdf{type}\;\var{v2}$}, $\{\var{v1}\})$ 
 $allow(${$\var{v1}\;\uri{p}\;\var{v2}$}, $\{\var{v1}\})$ 
 $include($$\var{v1}\;\urirdf{type}\;\uri{c}\rightarrow^{\exists}\var{v1}\;\uri{p}\;\var{v2}\;)$
 \end{lstlisting}
 \\ 
 \hline

 \begin{lstlisting}
 $\uri{s}\;\urirdf{type}\;\urish{NodeShape}$; 
   $\urish{targetObjectsOf}\;\uri{p}$$\,;$
   $\urish{property}$  [
       $\urish{path}\;\uri{q}$$\,;$
       $\urish{minCount}$ 1$\,;$ ]$\,.$
 \end{lstlisting}  &  \begin{lstlisting}
 $restrict(${$\var{v1}\;\uri{p}\;\var{v2}$}, $\{\var{v1}, \var{v2}\})$ 
 $allow(${$\var{v1}\;\uri{q}\;\var{v2}$}, $\{\var{v1}\})$ 
 $include($$\var{v1}\;\uri{p}\;\var{v2}\rightarrow^{\exists}\var{v2}\;\uri{q}\;\var{v3}\;)$
 \end{lstlisting}
 \\ 
 
 \hline
 \begin{lstlisting}
 $\uri{s}\;\urirdf{type}\;\urish{NodeShape}$; 
   $\urish{targetSubjectsOf}\;\uri{p}$$\,;$
   $\urish{property}$  [
       $\urish{path}\;\uri{q}$$\,;$
       $\urish{minCount}$ 1$\,;$ ]$\,.$
 \end{lstlisting}  &  \begin{lstlisting}
 $allow(${$\var{v1}\;\uri{p}\;\var{v2}$}, $\{v1\})$ 
 $allow(${$\var{v1}\;\uri{q}\;\var{v2}$}, $\{v1\})$ 
 $include($$\var{v1}\;\uri{p}\;\var{v2}\rightarrow^{\exists}\var{v1}\;\uri{q}\;\var{v3}\;)$
 \end{lstlisting}
 \\ 
 \hline

 \begin{lstlisting}
 $\uri{s}\;\urirdf{type}\;\urish{NodeShape}$$\,;$ 
   $\urish{targetClass}\;\uri{c}$$\,;$
   $\urish{property}$  [
       $\urish{path}\;[\urish{inversePath}\;\uri{p}]$$\,;$
       $\urish{minCount}$ 1$\,;$ ]$\,.$
 \end{lstlisting}  &  \begin{lstlisting}
 $allow(${$\var{v1}\;\urirdf{type}\;\var{v2}$}, $\{\var{v1}\})$ 
 $allow(${$\var{v1}\;\uri{p}\;\var{v2}$}, $\{\var{v1}\})$ 
 $include($$\var{v1}\;\urirdf{type}\;\uri{c}\rightarrow^{\exists}\var{v2}\;\uri{p}\;\var{v1}\;)$
 \end{lstlisting}
 \\ 
 \hline

 \begin{lstlisting}
 $\uri{s}\;\urirdf{type}\;\urish{NodeShape}$; 
   $\urish{targetObjectsOf}\;\uri{p}$$\,;$
   $\urish{property}$  [
       $\urish{path}\;[\urish{inversePath}\;\uri{q}]$$\,;$
       $\urish{minCount}$ 1$\,;$ ]$\,.$
 \end{lstlisting}  &  \begin{lstlisting}
 $restrict(${$\var{v1}\;\uri{p}\;\var{v2}$}, $\{\var{v1}, \var{v2}\})$ 
 $allow(${$\var{v1}\;\uri{q}\;\var{v2}$}, $\{\var{v1}\})$ 
 $include($$\var{v1}\;\uri{p}\;\var{v2}\rightarrow^{\exists}\var{v3}\;\uri{q}\;\var{v2}\;)$
 \end{lstlisting}
 \\ 
 
 \hline
 \begin{lstlisting}
 $\uri{s}\;\urirdf{type}\;\urish{NodeShape}$; 
   $\urish{targetSubjectsOf}\;\uri{p}$$\,;$
   $\urish{property}$  [
       $\urish{path}\;[\urish{inversePath}\;\uri{q}]$$\,;$
       $\urish{minCount}$ 1$\,;$ ]$\,.$
 \end{lstlisting}  &  \begin{lstlisting}
 $allow(${$\var{v1}\;\uri{p}\;\var{v2}$}, $\{v1\})$ 
 $allow(${$\var{v1}\;\uri{q}\;\var{v2}$}, $\{v1\})$ 
 $include($$\var{v1}\;\uri{p}\;\var{v2}\rightarrow^{\exists}\var{v3}\;\uri{q}\;\var{v1}\;)$
 \end{lstlisting}
 \\ 
 \hline


 \begin{lstlisting}
 $\uri{s}\;\urirdf{type}\;\urish{NodeShape}$$\,;$ 
   $\urish{targetClass}\;\uri{c}$$\,;$
   $\urish{property}$  [
       $\urish{path}\;\uri{p}$$\,;$
       $\urish{hasValue}\;\uri{a}\,;$
       $\urish{minCount}$ 1$\,;$ ]$\,.$
 \end{lstlisting}  &  \begin{lstlisting}
 $allow(${$\var{v1}\;\urirdf{type}\;\var{v2}$}, $\{\var{v1}\})$ 
 $allow(${$\var{v1}\;\uri{p}\;\var{v2}$}, $\{\var{v1}\})$ 
 $include($$\var{v1}\;\urirdf{type}\;\uri{c}\;\rightarrow^{\exists} \var{v1}\;\uri{p}\;\uri{a}\;)$
 \end{lstlisting}
 \\ 
 \hline

 \begin{lstlisting}
 $\uri{s}\;\urirdf{type}\;\urish{NodeShape}$; 
   $\urish{targetObjectsOf}\;\uri{p}$$\,;$
   $\urish{property}$  [
       $\urish{path}\;\uri{q}$$\,;$
       $\urish{hasValue}\;\uri{a}\,;$
       $\urish{minCount}$ 1$\,;$ ]$\,.$
 \end{lstlisting}  &  \begin{lstlisting}
 $restrict(${$\var{v1}\;\uri{p}\;\var{v2}$}, $\{\var{v1}, \var{v2}\})$ 
 $allow(${$\var{v1}\;\uri{q}\;\var{v2}$}, $\{\var{v1}\})$ 
 $include($$\var{v1}\;\uri{p}\;\var{v2}\rightarrow^{\exists}\var{v2}\;\uri{q}\;\uri{a}\;)$
 \end{lstlisting}
 \\ 
 
 \hline
 \begin{lstlisting}
 $\uri{s}\;\urirdf{type}\;\urish{NodeShape}$; 
   $\urish{targetSubjectsOf}\;\uri{p}$$\,;$
   $\urish{property}$  [
       $\urish{path}\;\uri{q}$$\,;$
       $\urish{hasValue}\;\uri{a}\,;$
       $\urish{minCount}$ 1$\,;$ ]$\,.$
 \end{lstlisting}  &  \begin{lstlisting}
 $allow(${$\var{v1}\;\uri{p}\;\var{v2}$}, $\{v1\})$ 
 $allow(${$\var{v1}\;\uri{q}\;\var{v2}$}, $\{v1\})$ 
 $include($$\var{v1}\;\uri{p}\;\var{v2}\rightarrow^{\exists}\var{v1}\;\uri{q}\;\uri{a}\;)$
 \end{lstlisting}
 \\ 
 \hline

 \begin{lstlisting}
 $\uri{s}\;\urirdf{type}\;\urish{NodeShape}$$\,;$ 
   $\urish{targetClass}\;\uri{c}$$\,;$
   $\urish{property}$  [
       $\urish{path}\;[\urish{inversePath}\;\uri{p}]$$\,;$
       $\urish{hasValue}\;\uri{a}\,;$
       $\urish{minCount}$ 1$\,;$ ]$\,.$
 \end{lstlisting}  &  \begin{lstlisting}
 $allow(${$\var{v1}\;\urirdf{type}\;\var{v2}$}, $\{\var{v1}\})$ 
 $allow(${$\var{v1}\;\uri{p}\;\var{v2}$}, $\{\var{v1}\})$ 
 $include($$\var{v1}\;\urirdf{type}\;\uri{c}\;\rightarrow^{\exists} \uri{a}\;\uri{p}\;\var{v1}\;)$
 \end{lstlisting}
 \\ 
 \hline

 \begin{lstlisting}
 $\uri{s}\;\urirdf{type}\;\urish{NodeShape}$; 
   $\urish{targetObjectsOf}\;\uri{p}$$\,;$
   $\urish{property}$  [
       $\urish{path}\;[\urish{inversePath}\;\uri{q}]$$\,;$
       $\urish{hasValue}\;\uri{a}\,;$
       $\urish{minCount}$ 1$\,;$ ]$\,.$
 \end{lstlisting}  &  \begin{lstlisting}
 $restrict(${$\var{v1}\;\uri{p}\;\var{v2}$}, $\{\var{v1}, \var{v2}\})$ 
 $allow(${$\var{v1}\;\uri{q}\;\var{v2}$}, $\{\var{v1}\})$ 
 $include($$\var{v1}\;\uri{p}\;\var{v2}\rightarrow^{\exists}\uri{a}\;\uri{q}\;\var{v2}\;)$
 \end{lstlisting}
 \\ 
 
 \hline
 \begin{lstlisting}
 $\uri{s}\;\urirdf{type}\;\urish{NodeShape}$; 
   $\urish{targetSubjectsOf}\;\uri{p}$$\,;$
   $\urish{property}$  [
       $\urish{path}\;[\urish{inversePath}\;\uri{q}]$$\,;$
       $\urish{hasValue}\;\uri{a}\,;$
       $\urish{minCount}$ 1$\,;$ ]$\,.$
 \end{lstlisting}  &  \begin{lstlisting}
 $allow(${$\var{v1}\;\uri{p}\;\var{v2}$}, $\{v1\})$ 
 $allow(${$\var{v1}\;\uri{q}\;\var{v2}$}, $\{v1\})$ 
 $include($$\var{v1}\;\uri{p}\;\var{v2}\rightarrow^{\exists}\uri{a}\;\uri{q}\;\var{v1}\;)$
 \end{lstlisting}
 \\ 
 \hline

 \hline
 \begin{lstlisting}
 $\uri{s}\;\urirdf{type}\;\urish{NodeShape}$; 
   $\urish{targetNode}\;\uri{a}$$\,;$
   $\urish{property}$  [
       $\urish{path}\;\uri{p}$$\,;$
       $\urish{node}\;\uri{k}\,;$]$\,.$
 $\uri{k}\;\urirdf{type}\;\urish{NodeShape}$; 
   $\urish{property}$  [
       $\urish{path}\;\uri{q}$$\,;$
       $\urish{hasValue}\;\uri{b}\,;$]$\,.$
 \end{lstlisting}  &  \begin{lstlisting}
 $allow(${$\var{v1}\;\uri{p}\;\var{v2}$}, $\{v1\})$ 
 $allow(${$\var{v1}\;\uri{q}\;\var{v2}$}, $\{v1\})$ 
 $include($$\uri{a}\;\uri{p}\;\var{v1}\rightarrow^{\exists}\var{v1}\;\uri{q}\;\uri{b}\;)$
 \end{lstlisting}
 \\ 
 \hline

 \hline
 \begin{lstlisting}
 $\uri{s}\;\urirdf{type}\;\urish{NodeShape}$; 
   $\urish{targetNode}\;\uri{a}$$\,;$
   $\urish{property}$  [
       $\urish{path}\;$
           $[\urish{inversePath}\;\uri{p}]$$\,;$
       $\urish{node}\;\uri{k}\,;$]$\,.$
 $\uri{k}\;\urirdf{type}\;\urish{NodeShape}$; 
   $\urish{property}$  [
       $\urish{path}\;\uri{q}$$\,;$
       $\urish{hasValue}\;\uri{b}\,;$]$\,.$
 \end{lstlisting}  &  \begin{lstlisting}
 $allow(${$\var{v1}\;\uri{p}\;\var{v2}$}, $\{v1\})$ 
 $allow(${$\var{v1}\;\uri{q}\;\var{v2}$}, $\{v1\})$ 
 $include($$\var{v1}\;\uri{p}\;\uri{a}\rightarrow^{\exists}\var{v1}\;\uri{q}\;\uri{b}\;)$
 \end{lstlisting}
 \\ 
 \hline

 \hline
 \begin{lstlisting}
 $\uri{s}\;\urirdf{type}\;\urish{NodeShape}$; 
   $\urish{targetNode}\;\uri{a}$$\,;$
   $\urish{property}$  [
       $\urish{path}\;\uri{p}$$\,;$
       $\urish{node}\;\uri{k}\,;$]$\,.$
 $\uri{k}\;\urirdf{type}\;\urish{NodeShape}$; 
   $\urish{property}$  [
       $\urish{path}\;$
           $[\urish{inversePath}\;\uri{q}]$$\,;$
       $\urish{hasValue}\;\uri{b}\,;$]$\,.$
 \end{lstlisting}  &  \begin{lstlisting}
 $allow(${$\var{v1}\;\uri{p}\;\var{v2}$}, $\{v1\})$ 
 $allow(${$\var{v1}\;\uri{q}\;\var{v2}$}, $\{v1\})$ 
 $include($$\uri{a}\;\uri{p}\;\var{v1}\rightarrow^{\exists}\uri{b}\;\uri{q}\;\var{v1}\;)$
 \end{lstlisting}
 \\ 
 \hline

 \hline
 \begin{lstlisting}
 $\uri{s}\;\urirdf{type}\;\urish{NodeShape}$; 
   $\urish{targetNode}\;\uri{a}$$\,;$
   $\urish{property}$  [
       $\urish{path}\;$
           $[\urish{inversePath}\;\uri{p}]$$\,;$
       $\urish{node}\;\uri{k}\,;$]$\,.$
 $\uri{k}\;\urirdf{type}\;\urish{NodeShape}$; 
   $\urish{property}$  [
       $\urish{path}\;$
           $[\urish{inversePath}\;\uri{q}]$$\,;$
       $\urish{hasValue}\;\uri{b}\,;$]$\,.$
 \end{lstlisting}  &  \begin{lstlisting}
 $allow(${$\var{v1}\;\uri{p}\;\var{v2}$}, $\{v1\})$ 
 $allow(${$\var{v1}\;\uri{q}\;\var{v2}$}, $\{v1\})$ 
 $include($$\var{v1}\;\uri{p}\;\uri{a}\rightarrow^{\exists}\uri{b}\;\uri{q}\;\var{v1}\;)$
 \end{lstlisting}
 \\ 
 \hline
 
  \hline
 \begin{lstlisting}
 $\uri{s}\;\urirdf{type}\;\urish{NodeShape}$; 
   $\urish{targetNode}\;\uri{a}$$\,;$
   $\urish{or}\;($
     [ $\urish{property}$ [
       $\urish{path}\;\uri{p}$$\,;$
       $\urish{node}\;\uri{k}$$\,;$ ] 
     ]
     [ $\urish{property}$ [
       $\urish{path}\;\uri{p}$$\,;$
       $\urish{node}\;$ [ 
         $\urish{not}\;$ [ 
           $\urish{in}\;$ ($\uri{b}$ ) ] ] $\,;$ ]
     ] ) .
 $\uri{k}\;\urirdf{type}\;\urish{NodeShape}$; 
   $\urish{property}$  [
     $\urish{path}\;$
       [ $\urish{inversePath}\;\uri{q}$ ] $\,;$
     $\urish{minCount}\;$ 1 $\,;$]$\,.$
 \end{lstlisting}  &  \begin{lstlisting}
 $allow(${$\var{v1}\;\uri{p}\;\var{v2}$}, $\{v1\})$ 
 $allow(${$\var{v1}\;\uri{q}\;\var{v2}$}, $\{v1\})$ 
 $include($$\uri{a}\;\uri{p}\;\uri{b}\rightarrow^{\exists}\var{v1}\;\uri{q}\;\uri{b}\;)$
 \end{lstlisting}
 \\ 
 \hline
 
   \hline
 \begin{lstlisting}
 $\uri{s}\;\urirdf{type}\;\urish{NodeShape}$; 
   $\urish{targetNode}\;\uri{a}$$\,;$
   $\urish{or}\;($
     [ $\urish{property}$ [
       $\urish{path}\;\uri{p}$$\,;$
       $\urish{node}\;\uri{k}$$\,;$ ] 
     ]
     [ $\urish{property}$ [
       $\urish{path}\;\uri{p}$$\,;$
       $\urish{node}\;$ [ 
         $\urish{not}\;$ [ 
           $\urish{in}\;$ ($\uri{b}$ ) ] ] $\,;$ ]
     ] ) .
 $\uri{k}\;\urirdf{type}\;\urish{NodeShape}$; 
   $\urish{property}$  [
     $\urish{path}\;$
       [ $\urish{inversePath}\;\uri{q}$ ] $\,;$
     $\urish{hasValue}\;\uri{c}\,;$]$\,.$
 \end{lstlisting}  &  \begin{lstlisting}
 $allow(${$\var{v1}\;\uri{p}\;\var{v2}$}, $\{v1\})$ 
 $allow(${$\var{v1}\;\uri{q}\;\var{v2}$}, $\{v1\})$ 
 $include($$\uri{a}\;\uri{p}\;\uri{b}\rightarrow^{\exists}\uri{c}\;\uri{q}\;\uri{b}\;)$
 \end{lstlisting}
 \\ 
 \hline

 \end{longtable}
 \begin{table}
  \vspace{-\baselineskip}
   \vspace{-\baselineskip}
    \vspace{-\baselineskip}
     \vspace{-\baselineskip}  \vspace{-\baselineskip}
   \vspace{-\baselineskip}
    \vspace{-\baselineskip}
     \vspace{-\baselineskip}
       \vspace{-\baselineskip}
   \vspace{-\baselineskip}
    \vspace{-\baselineskip}
     \vspace{-\baselineskip}
 \caption{Templates of SHACL shapes and their translation transformations. To reduce the size of this table we have omitted some templates that can be obtained from the ones above by simply switching the subject and object of triple patterns.} \label{tab:transf}

  \normalsize
 \end{table}
 \end{center}
 
 \pagebreak

\FloatBarrier

\subsection{Translation form Triplestore Schemas to SHACL constraints}

Given a schema $S: \langle S^{G}, S^{\Delta}, S^{\exists} \rangle$ we reconstruct a corresponding set of SHACL shapes in two phases. In the first phase, we use Table \ref{tab:transf} in the opposite direction than in the previous section, to deal with existential rules. More specifically, we match each existential constraint in $S^{\exists}$ with the existential constraints of Table \ref{tab:transf}, and re-create the corresponding SHACL shape.

Unfortunately we cannot use the table in a similar fashion to translate the graph pattern and no literal sets in $S$, as $S^{G}$ and $S^{\Delta}$ might not correspond to a single template. Instead, they might be a result of the interactions of multiple such templates. Therefore to compute the corresponding SHACL shapes of $S^{G}$ and $S^{\Delta}$  we present Algorithm \ref{algComplete} and its helper algorithms \ref{algAddConstants} and \ref{algAddConstants2}. Algorithm \ref{algComplete} computes the \texttt{SHACL} graph for a schema $S$ on a predicate by predicate basis. The SHACL translation of graph pattern $S^{G}$ and no literal set $S^{\Delta}$ can then be computed by executing Algorithm \ref{algComplete} for each predicate $p$ in $S^G$.  
In the algorithms we use notation $e \ll e'$ to indicate that an element $e$ is subsumed by another element $e'$. This condition is true if and only if one of these conditions apply: (1) $e = e'$ (2) $e'$ is a variable that allows for literals or (3) $e$ is a \texttt{URI} and $e'$ is a variable.

\begin{algorithm}[H]
\footnotesize
    \caption{Helper algorithm to convert a set of elements $e$ into \texttt{SHACL} constraints.}
    \label{algAddConstants}
    \begin{algorithmic}[1]
        \Procedure{compute\_constraints}{$e, S^{\Delta}$} 
        \State $r \leftarrow$ empty set of triples
        \State $b, b' \leftarrow  $ a new fresh \texttt{URI} each
            \If{$e$ contains a variable not in the no-literal set}
            \State{$\texttt{add} \; \langle b,\urish{nodeKind},\urish{IRIOrLiteral} \rangle $ to $r$}
            \Else
                 \If{$e$ contains a variable in the no-literal set}
                    \State{$\texttt{add} \; \langle b,\urish{nodeKind},\urish{IRI} \rangle $ to $r$}
                \EndIf
                \If{$e$ contains constants}
                  \State{$s \leftarrow $ \texttt{SHACL} list of constants in $e$} \State{$\texttt{add} \; \langle b',\urish{in},s \rangle $ to $r$}
                \EndIf
            \EndIf
        \Return $r$
        \EndProcedure
    \end{algorithmic}
\end{algorithm}

\begin{algorithm}[H]
\footnotesize
    \caption{Helper algorithms to add constraints $C$ to a shape $i$}
    \label{algAddConstants2}
    \begin{algorithmic}[1]
        \Procedure{add\_constraints}{$i,C$} 
        \State $r \leftarrow$ empty set of triples
        \State $B \leftarrow $ \texttt{SHACL} list of $\{t[1] | t \in C\} $
        \State $\texttt{add} \; \langle i,\urish{or},B \rangle $ to $r$
        
        \Return $r$
        \EndProcedure
    \end{algorithmic}
\end{algorithm}

\begin{algorithm}
\footnotesize
    \caption{Compute the SHACL triples for predicate $p$ and schema $\langle S^G,S^{\Delta} \rangle$}
    \label{algComplete}
    \begin{algorithmic}[1]
        \Procedure{computeShape}{$p$,$S^G$,$S^{\Delta}$} 
            \State $r \leftarrow$ empty set of triples
            \For{each $t \in S^G$ s.t.\ $t[1]$ or $t[3]$ is a constant}
                \State $i \leftarrow  $ a new fresh \texttt{URI}
                \State $\texttt{add} \; \langle i, \urirdf{type}, \urish{NodeShape} \rangle $ to $r$
                \If{$t[1]$ is a \texttt{URI}}
                    \State $\texttt{add} \; \langle i,\urish{targetNode},t[1] \rangle $ and $\langle i,\urish{path},p \rangle $ to $r$
                    \State $e \leftarrow \{t'[3] | \forall t' \in S^G \text{s.t.} \; t[1] \ll t'[1]\}$ 
                    \State $E \leftarrow $\Call{compute\_constraints}{$e, S^{\Delta}$}
                    \State add \Call{add\_constraints}{$i,E$} to $r$
                \EndIf
            \EndFor
            \If{$\not \exists t \in S^G \text{s.t.} \; t[1]$ is a variable}
                \State $i \leftarrow  $ a new fresh \texttt{URI}
                \State $\texttt{add} \; \langle i, \urirdf{type}, \urish{NodeShape} \rangle $ and $\langle i,\urish{targetSubjectsOf},p \rangle$ to $r$
                \State $e \leftarrow \{t'[1] | \forall t' \in S^G \text{s.t.} \; t'[1] \text{is constant}\}$
                \State $E \leftarrow $\Call{compute\_constraints}{$e, S^{\Delta}$}
                \State add \Call{add\_constraints}{$i,E$}   to $r$
            \Else
                \State $i, i' \leftarrow  $ a new fresh \texttt{URI} each
            \State $\texttt{add} \; \langle i, \urirdf{type}, \urish{NodeShape} \rangle $ and $\langle i, \urish{targetObjectsOf}, p \rangle$ to $r$
            \State $C \leftarrow \{t[3] | t \in S^G \text{s.t.} \; t[1] \text{is a variable}\}$ 
            \State $E \leftarrow $\Call{compute\_constraints}{$C, S^{\Delta}$}
            \For{each $t \in S^G$ s.t.\ $t[1]$ is a constant}
                \For{each $t' \in S^G$ s.t.\ $t'[3] \ll t[3]$}
                    \State $D \leftarrow \{t''[1] | t'' \in S^G \text{s.t.} \; t'[3] \ll t''[3]\}$
                    \State $b, b', b'' \leftarrow  $ a new fresh \texttt{URI} each
                    \State \texttt{add}  $\langle b, \urirdf{type}, \urish{NodeShape} \rangle $, $\langle b, \urish{property}, b' \rangle $,  
                    \Statex \hspace{58pt} $\langle b', \urish{path}, b'' \rangle $ and $\langle b'', \urish{inversePath}, p \rangle $ to $r$
                    \State $H \leftarrow $ \Call{compute\_constraints}{$D, S^{\Delta}$}
                    \State add \Call{add\_constraints}{$b',H$}  to $r$
                    
                    \State $n \leftarrow$ the only triple in  \Call{compute\_constraints}{$\{t'[3]\}, S^{\Delta}$}
                    \State \texttt{add}  $n$ and $\langle n[1], \urish{node}, b \rangle$ to $E$
                \EndFor
            \EndFor
            \State add \Call{add\_constraints}{$i,E$}  to $r$
            \State $E' \leftarrow $ empty set of triples
            \State $\texttt{add} \; \langle i', \urirdf{type}, \urish{NodeShape} \rangle $ and $\langle i', \urish{targetSubjectsOf}, p \rangle$ to $r$
            \For{each $t \in S^G$ s.t.\ $t[1]$ or $t[3]$ is a constant}
                \State $n \leftarrow$ the only triple in  \Call{compute\_constraints}{$\{t'[1]\}, S^{\Delta}$}
                \State add $n$ to $E'$
                \If{$t[3]$ is not a variable $v$ s.t.\ $v \not \in s^{\Delta}$}
                    \State $b, b' \leftarrow  $ a new fresh \texttt{URI} each
                    
                    \State \texttt{add}  $\langle b, \urirdf{type}, \urish{NodeShape} \rangle $, $\langle b, \urish{property}, b' \rangle $ and
                    \Statex \hspace{58pt} $\langle b', \urish{path}, p \rangle $ to $r$
                    \State$H \leftarrow $ empty set of triples
                    \If{$t[3]$ is a variable}
                        \State $H \leftarrow $ \Call{compute\_constraints}{$\{t[3]\}, S^{\Delta}$}
                    \Else
                        \State $b'' \leftarrow  $ a new fresh \texttt{URI}
                        \State $H \leftarrow  \{\langle b'', \urish{hasValue}, t[3] \rangle\}$
                    \EndIf
                    
                    \State add \Call{add\_constraints}{$b',H$}  to $r$

                    \State \texttt{add}  $\langle n[1], \urish{node}, b \rangle$ to $E'$
                \EndIf
            \EndFor
            \State add \Call{add\_constraints}{$i',E'$}  to $r$
            \EndIf
        \Return $r$
        \EndProcedure
    \end{algorithmic}
\end{algorithm}

\end{document}